\documentclass[journal]{IEEEtran}

\pdfoutput=1

\usepackage{graphicx}
\usepackage{multirow}
\usepackage{subfig,comment}
\usepackage{amsmath}
\usepackage[group-separator={,}]{siunitx}
\usepackage{cite}
\usepackage{hyperref}

\def \newadditions#1{\textit{#1}}
\def \newadditions#1{{#1}}

\begin{document}

\title{Surprise Search for Evolutionary Divergence}

\author{Daniele~Gravina, Antonios~Liapis, and~Georgios~N. Yannakakis
\thanks{All authors are with the Institute of Digital Games, University of Malta, Msida 2080, Malta 
(e-mail: daniele.gravina@um.edu.mt; antonios.liapis@um.edu.mt; georgios.yannakakis@um.edu.mt)
}
\thanks{
}}

\maketitle

\begin{abstract}
Inspired by the notion of \emph{surprise} for unconventional discovery we introduce a general search algorithm we name \emph{surprise search} as a new method of evolutionary divergent search. Surprise search is grounded in the divergent search paradigm and is fabricated within the principles of evolutionary search. The algorithm mimics the self-surprise cognitive process and equips evolutionary search with the ability to seek for solutions that deviate from the algorithm\rq{}s expected behaviour. The predictive model of expected solutions is based on historical trails of where the search has been and local information about the search space. Surprise search is tested extensively in a robot maze navigation task: experiments are held in four authored deceptive mazes and in 60 generated mazes and compared against objective-based evolutionary search and novelty search. The key findings of this study reveal that surprise search is advantageous compared to the other two search processes. In particular, it outperforms objective search and it is as efficient as novelty search in all tasks examined. Most importantly, surprise search is faster, on average, and more robust in solving the navigation problem compared to any other algorithm examined. Finally, our analysis reveals that surprise search explores the behavioural space more extensively and yields higher population diversity compared to novelty search. What distinguishes surprise search from other forms of divergent search, such as the search for novelty, is its ability to diverge not from earlier and seen solutions but rather from predicted and unseen points in the domain considered.
\end{abstract}


\IEEEpeerreviewmaketitle

\section{Introduction}\label{sec:introduction}

\IEEEPARstart{O}{ver} the last 50 years, evolutionary computation (EC) has shown vast potential in numerical and behavioural optimization. The most common approach to optimization in artificial evolution is via an \emph{objective function}, which rewards solutions based on their `goodness' \cite{goldberg1988genetic}, i.e. how close they are to an optimal behaviour (if such a behaviour is known beforehand) or how much they improve a performance metric. The objective function (or fitness function) encapsulates the principle of evolutionary pressure for fitting (adapting) within the environment. Despite the success of such approaches in a multitude of tasks \cite{goldberg1988genetic,michalski2013machine}, they are challenged in deceptive fitness landscapes \cite{goldberg1987deceptive} where the global optimum is neighboured by low-quality solutions. In such cases, the local search of an objective-based evolutionary algorithm can guide search away from a global optimum and towards local optima. As a general principle, more deceptive problems challenge the design of a corresponding objective function; this paper follows \cite{whitley1991deception} and views deception as the \emph{intuitive definition of problem hardness}. Many algorithms have been proposed to tackle the problem of deception, primarily revolving around diversity preservation \cite{goldberg1987sharing,hu2005hierarchical,hornby2006alps}, which deters premature convergence while still rewarding proximity to the objective and divergent search \cite{lehman2011abandoning} which abandons objectives in favour of rewarding diversity in the population.

\newadditions{
There are, however, problems which lack an easily defined objective --- or a gradient to reaching it. For instance, open-ended evolution studies within artificial life \cite{channon2001passing} do not have a goal state and instead prioritize e.g. survival \cite{yaeger1994poly,adami2000evolution}. In evolutionary art, music or design, a large body of research in computational creativity and generative systems \cite{boden2004creativemind,ritchie2007some,wiggins2006preliminary} focuses on the \emph{creative} capacity of search rather than on the objectives. In \cite{ritchie2007some}, computational creativity is considered on two core properties of a produced solution: value and novelty \cite{ritchie2007some}. Value is the degree to which a solution is of high quality, whereas novelty is the degree to which a solution (or output) is dissimilar to existing examples.
While objective-based EC can be seen as a metaphor of searching for value, a divergent EC strategy as \emph{novelty search} \cite{lehman2011abandoning,lehman2013diversity} can be considered as a metaphor of searching for novelty.
An effective use of both in EC can lead to highly novel and valuable at the same time outcomes \cite{liapis2013delenox}, thus realizing \emph{quality diversity} \cite{pugh2016quality}. However, according to \cite{grace2015modeling}, novelty and value are not sufficient for the discovery of highly creative and unconventional solutions to problems. While novelty can be considered as a static property, surprise considers the temporal properties of the discovery, an important dimension to assess the creativity of the generated solution \cite{grace2015modeling,maher2010evaluating}. Further studies in general intelligence and decision making \cite{barto2013novelty} support the importance of unexpectedness for problem-solving.
}

Driven by the notion of computational surprise for the purpose of creatively traversing the search space towards unexpected or serendipitous solutions, this paper proposes \emph{surprise search} for the purposes of divergent evolutionary search. The hypothesis is that searching for unexpected --- not merely unseen --- solutions is beneficial to EC as it complements our search capacities with highly efficient and robust algorithms beyond the search for objectives or mere novelty. Surprise search is built upon the novelty search \cite{lehman2011abandoning} paradigm that rewards individuals which differ from other solutions in the same population and a historical archive. 
Surprise is assumed to arise from a violation of expectations \cite{lorini2007cognitive}: as such, it is different than novelty which rewards deviation from past and current behaviours. A computational, quantifiable model of surprise must build expectations based on trends in past behaviours, and predict future behaviours from which it must diverge from. In order to create expected behaviours, the algorithm maintains a lineage of where evolutionary search has been. These groups of evolutionary lineages require the right level of locality in the behavioural space --- surprise can be inclusive of all behaviours (globally) or merely consider part of all possible behaviours (locally).  
Any deviation from these stepping stones of search would elicit surprise; alternatively, they can be viewed as serendipitous discovery if the deviation leads to a surprisingly good point in the behavioural space. The findings of this paper suggest that surprise constitutes a powerful drive for computational discovery as it incorporates predictions of an expected behaviour that it attempts to deviate from; these predictions may be based on behavioural relationships in the solution space as well as historical trends derived from the algorithm\rq{}s sampling of the domain.

This paper builds upon and extends significantly our earlier work which introduced the notion of surprise search \cite{yannakakis2016searching} and compared its performance against novelty search and objective-based search \cite{gravina2016surprisebeyond} in the maze navigation testbed of \cite{lehman2011abandoning}. The current paper extends the preliminary study of \cite{gravina2016surprisebeyond} by introducing two new maze navigation problems of increased complexity, and analysing the impact of several parameters to the behaviour of surprise search. This paper also includes an extensive comparison between novelty search and surprise search both in the behavioural and the genotypic space. Finally, to further examine how surprise search performs across a wide range of problems, we test how it scales in sixty randomly generated mazes of varying degrees of complexity.
The key findings of the paper suggest that surprise search is as efficient as novelty search and both algorithms, unsurprisingly, outperform fitness-based search. Furthermore, surprise search appears to be the most robust algorithm in the four testbed tasks and to be the most successful and fastest algorithm in the 60 randomly generated mazes. While both novelty and surprise search converge to the solution significantly faster than fitness-based search, surprise search solves the navigation problem faster, on average, and more often than novelty search. The experiments of this paper validate our hypothesis that surprise can be beneficial as a divergent search approach and provide evidence for its supremacy over novelty search in the tasks examined.

\section{Deception, Divergent Search and Quality Diversity}\label{sec:relatedwork}

This section motivates surprise search by providing a brief overview of the challenges faced by fitness-based approaches when handling \emph{deceptive} problems and how \emph{divergent search} has been used to address such challenges. The section concludes with a discussion on the relationship between surprise search and \emph{quality diversity} algorithms. 

\subsection{Deception in Evolutionary Computation}\label{sec:relatedwork_deception}

The term \emph{deception} in the context of EC was introduced by \cite{goldberg1987deceptive} to describe instances where highly-fit building blocks, when recombined, may guide search away from the global optimum. Since that first mention, the notion of deception (including deceptive problems and deceptive search spaces) has been refined and expanded to describe several problems that challenge evolutionary search for a solution. \cite{whitley1991deception} argues that ``the only challenging problems are deceptive''. EC-hardness is often attributed to deception, as well as sampling error \cite{liepins1990representational} and a rugged fitness landscape \cite{kauffman1989rugged}. In combinatorial optimisation problems, the fitness landscape can affect optimisation when performing local search. Such a search process assumes that there is a high correlation between the fitness of neighbouring points in the search space, and that genes in the chromosome are independent of each other. The latter assumption refers to \emph{epistasis} \cite{davidor1991epistasis} which is a factor of GA-hardness: when epistasis is high (i.e. where too many genes are dependent on other genes in the chromosome), the algorithm searches for a unique optimal combination of genes but no substantial fitness improvements are noticed \cite{davidor1991epistasis}. 

As noted, epistasis is evaluated from the perspective of the fitness function and thus is susceptible to deception; \cite{naudts1999epistasis} argue that deceptive functions can not have low epistasis, although fitness functions with high epistasis are not necessarily deceptive. Such approaches are often based on the concepts of correlation, i.e. the degree to which an individual's fitness score is well correlated to its neighbours' in the search space, or epistasis, i.e. the degree of interaction among genes' effects. As noted by \cite{lehman2011abandoning}, most of the factors of EC-hardness originate from the fitness function itself; however, poorly designed genetic operators and poorly chosen evolutionary parameters can exacerbate the problem.

\subsection{Divergent Search}\label{sec:relatedwork_divergent}

\newadditions{
By definition, the biggest issue of a deceptive problem is premature convergence to local optima, while the global optimum is difficult to reach as the deceptive fitness landscape lead the search away from it.
Several approaches have been proposed to counter this behaviour, surveyed by \cite{lehman2013diversity}. For instance, speciation \cite{stanley2002neat} and niching \cite{wessing2013niching} are popular diversity maintenance techniques, which enforce local competition among similar solutions.  Similarity can be measured on the genotypical level \cite{goldberg1987sharing}, on the fitness scores \cite{hu2005hierarchical}, or on the age of the individuals \cite{hornby2006alps}.
Multiple promising directions are favoured by this localised competition, making premature convergence less likely.
}
Alternatives such as coevolution \cite{angeline1994competitive} can lead to an arms race that finds a better gradient for search, but can suffer when individuals' performance is poor or one individual vastly outperforms others \cite{ficici1998stablemates}. 
Finally, techniques from multi-objective optimisation can, at least in theory, explore the search space more effectively by evaluating individuals in more than one measures of quality \cite{knowles2001optima} and thus avoid local optima in one fitness function by attempting to improve other objectives; however, multi-objective optimisation can not guarantee to circumvent deception \cite{deb1999multiobjective}.

\newadditions{
Divergent search methods differ from previous approaches as they explicitly ignore the objective of the problem they are trying to solve. While the approaches described above provide control mechanisms, modifiers or alternate objectives which complement the gradient search towards a better solution, divergent algorithms such as novelty search \cite{lehman2011abandoning} motivates exploration of the search space by rewarding individuals that are phenotypically (or behaviourally) different without considering whether they are objectively ``better'' than others. Novelty search is neither random walk nor exhaustive search, however, as it gives higher rewards to solutions that are different from others in the current and past populations by maintaining a memory of the areas of the search space it has explored via a \emph{novelty archive}. The archive contains previously found novel individuals, and the highest-scoring individuals in terms of novelty are continuously added to it as evolution carries on. The distance measure which assesses ``difference'' is based on a behaviour characterization, which is problem-dependent: for a maze navigation task, distance may be calculated on the agents' final positions or directions \cite{lehman2011abandoning, pugh2016quality}, for robot locomotion it may be on the position of a robot's centre of mass \cite{lehman2011abandoning}, for evolutionary art it may be on properties of the images such as brightness and symmetry \cite{lehman2012impressiveness}.
}

\subsection{Quality Diversity}

A recent trend in evolutionary computation is inspired by a species\rq{} tendency to face a strong competition for survival within its own niche \cite{pugh2016quality}. EC algorithms of this type seek for the discovery of both \emph{quality} and \emph{diversity} at the same time, following the traditional approach within computational creativity of seeking outcomes characterized by both quality (or \emph{value}) and novelty \cite{ritchie2007some}. Such evolutionary algorithms have been named \emph{quality diversity} algorithms \cite{pugh2016quality} and aim to find a maximally diverse population of highly performing individuals. Examples of such algorithms include novelty search with local competition \cite{lehman2011evolving} and MAP-Elites \cite{mouret2015illuminating} as well as algorithms that constrain the feasible space of solutions --- thereby forcing high quality solutions --- while searching for divergence such as \emph{constrained novelty search} \cite{liapis2015ecj,liapis2013delenox}. 

Surprise search can complement the search for diversity and replace other divergent search algorithms commonly used for quality diversity. It can, for instance, be used in combination with local competition instead of novelty search as in \cite{lehman2011evolving}.
Towards that goal, a recent study employed surprise search for game weapon generation in a constrained fashion \cite{gravina2016constrained}. The \emph{constrained surprise search} algorithm rewarded the generation of surprising weapons --- thereby maintaining diversity --- with guaranteed high quality imposed by constraints on weapon balance and effectiveness.

\section{The Notion of Surprise Search}\label{sec:surprise}

\newadditions{
This section examines the notion of surprise as a driver of evolutionary search. To this end, we first describe the concept of surprise, we then highlight the differences between surprise and novelty and, finally, we frame surprise search in the context of divergent search.
}

\subsection{What is Surprise?}\label{sec:surprise_whatis}

The study of surprise has been central in neuroscience, psychology, cognitive science, and to a lesser degree in computational creativity and computational search. In psychology and emotive modelling studies, surprise defines one of Ekman\rq{}s six basic emotions~\cite{ekman1992argument}. Within cognitive science, surprise has been defined as a temporal-based cognitive process of the unexpected~\cite{meyer1997toward,lorini2007cognitive}, a violation of a belief~\cite{ortony1987surprisingness}, a reaction to a mismatch~\cite{lorini2007cognitive}, or a response to novelty~\cite{wiggins2006preliminary}. In computational creativity, surprise has been attributed to the creative output of a computational process~\cite{grace2015modeling,wiggins2006preliminary}.

While variant types and taxonomies of surprise have been suggested in the literature --- such as aggressive versus passive surprise \cite{grace2015modeling} --- we can safely derive a definition of surprise that is general across all disciplines that study surprise as a phenomenon. For the purposes of this paper we define surprise as the \emph{deviation from the expected} and we use the notions \emph{surprise} and \emph{unexpectedness} interchangeably due to their highly interwoven nature. 

\subsection{Novelty vs. Surprise}\label{sec:surprise_versus_theory}

Novelty and surprise are different notions by definition as it is possible for a solution to be both novel and/or expected to variant degrees. Following the core principles of Lehman and Stanley \cite{lehman2011abandoning} and Grace et al. \cite{grace2015modeling}, novelty is defined as the degree to which a solution is \emph{different from prior} solutions to a particular problem. On the other hand, surprise is the degree to which a solution is \emph{different from the expected} solution to a particular problem.

Expectations are naturally based on inference from past experiences; analogously surprise is built on the temporal model of past behaviours. To exemplify the difference between the notions of novelty and surprise, \cite{yannakakis2016searching} uses a card memory game where cards are revealed, one at a time, to the player who has to predict which card will be revealed next. The novelty of game outcome (i.e. next card) is the highest if all past revealed cards are different. The surprise value of the game outcome in that case is low as the player has grown to expect a new, unseen, card every time. On the other hand, if seen cards are revealed after a while then the novelty of the game outcome decreases, but surprise increases as the game deviates from the expected behaviour which calls for a new card every time. 

Surprise is a temporal notion as expectations are by nature temporal. Prior information is required to predict what is expected; hence a \emph{prediction of the expected} \cite{maher2010evaluating} is a necessary component for modelling surprise computationally. By that logic, surprise can be viewed as a \emph{temporal novelty} process or as novelty on the prediction (rather than the behavioural) space. 
Surprise search maintains a prediction (a gradient) of where novelty has been on the prediction space, which is derived from the behavioural space. In that sense surprise resembles a \emph{time derivative} of novelty.

\begin{figure}[!tb]
\centering
\includegraphics[width=0.48\textwidth]{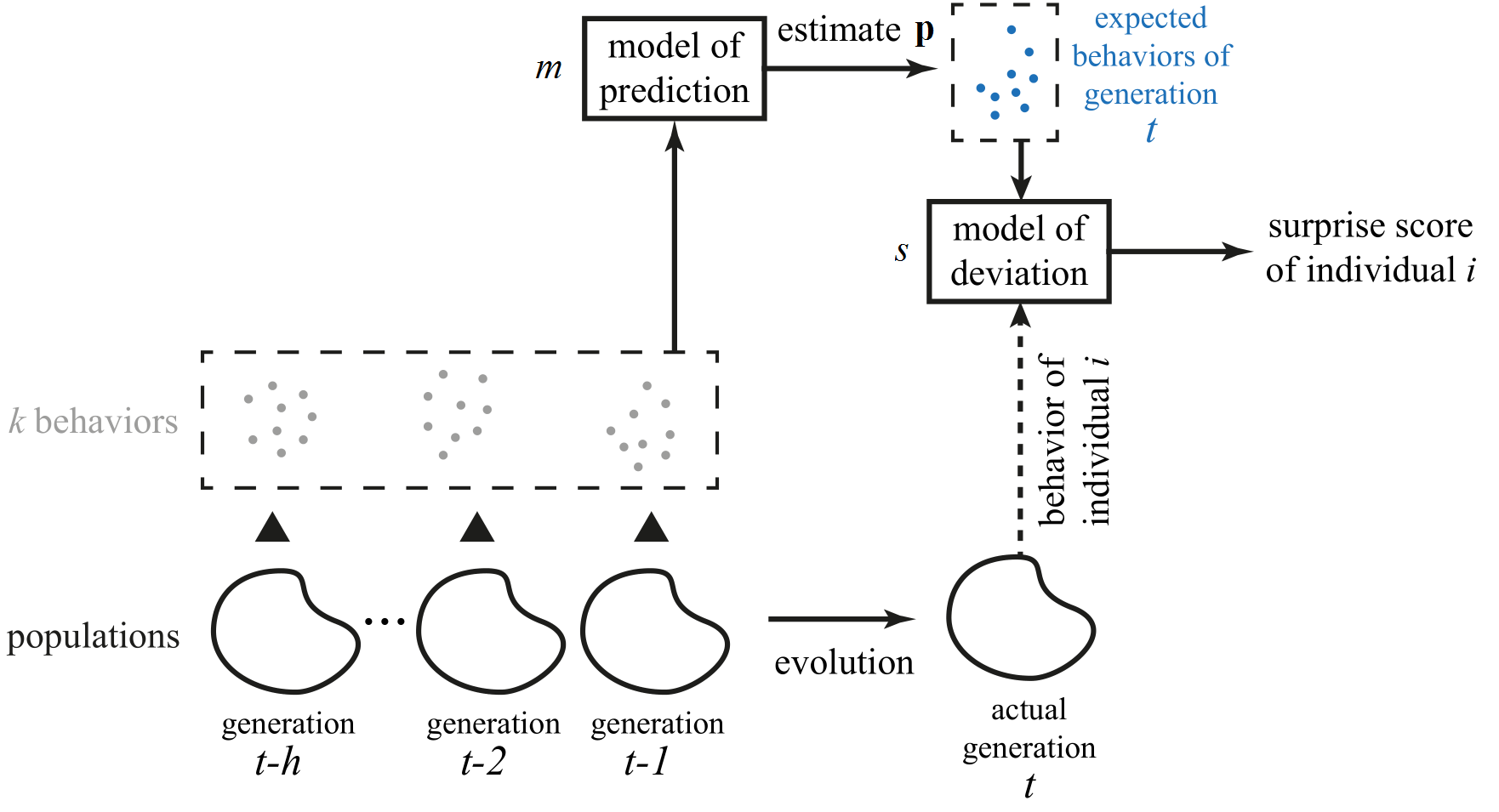}
\caption{High-level overview of the surprise search algorithm when evaluating an individual $i$ in a population at generation $t$. The $h$ previous generations are considered, with respect to $k$ behavioural characteristics per generation, to predict the expected $k$ behaviours of generation $t$. The surprise score of individual $i$ is the deviation of the behaviour of $i$ from a subset of these $k$ expected behaviours.}
\label{fig:surpriseUML}
\end{figure}

\subsection{Novelty Search vs. Surprise Search}\label{sec:surprise__versus_search}

According to Grace et al. \cite{grace2015modeling}, novelty and value (i.e. objective in the context of EC) are not sufficient for the discovery of unconventional solutions to problems (or creative outputs) as novelty does not cater for the temporal aspects of discovery. Novelty search rewards divergence from \emph{current} behaviours (i.e.~other individuals of the population) and \emph{prior} behaviours (i.e.~an archive of previously novel solutions) \cite{lehman2011abandoning}; in this way it provides the necessary stepping stones toward achieving an objective (i.e. value). Surprise, on the other hand, complements the search for novelty as it rewards divergence from the \emph{expected} behaviour. In other words while novelty search attempts to deviate from previously seen solutions, surprise search attempts to deviate from solutions that are expected to be seen in the future. 

Highly relevant to this study is the work on computational models of surprise for artificial agents \cite{macedo2001modeling}, which however does not consider using such a model of surprise for search. Other aspects of unexpectedness such as intrinsic motivation \cite{oudeyer2007intrinsic} and artificial curiosity \cite{schmidhuber2010formal} have also been modelled. The concepts of novelty within reinforcement learning research are also interlinked to the idea of surprise search \cite{kaplan2006information,oudeyer2007intrinsic}. Artificial curiosity and intrinsic motivation differ from surprise search as the latter is based on evolutionary divergent search and motivated by open-ended evolution, similarly to novelty search. Specifically, surprise search does not keep a persistent world model as \cite{schmidhuber2010formal} does; instead it focuses on the current trajectory of search using the latest points of the search space it has explored (and ordering them temporally). Additionally, it rewards deviations from expected behaviours agnostically rather than based on how those deviations improve a world model. This allows surprise search to backtrack and re-visit areas of the search space it has already visited, which is discouraged in both novelty search and curiosity.

Inspired by the above arguments and findings in computational creativity, we view surprise for computational search as the degree to which expectations about a solution are violated through observation \cite{grace2015modeling}. Our hypothesis is that if modelled appropriately, surprise may enhance divergent search and complement or even surpass the performance of traditional forms of divergent search such as novelty. The main findings of this paper validate our hypothesis.

\section{The Surprise Search Algorithm}\label{sec:surprisesearch}

This section discusses the principles of designing a surprise search algorithm for any task or search space. To realise surprise as a search mechanism, an individual should be rewarded when it \emph{deviates from the expected behaviour}, i.e. the evaluation of a population in evolutionary search is adapted. This means that surprise search can be applied to any EC method, such as NEAT \cite{stanley2002neat} in the case study of this paper.

As discussed above, surprise search must reward an individual's deviation from the expected behaviour. This goal can be decomposed into two tasks: \emph{prediction} and \emph{deviation}. At the highest descriptive level, surprise search uses local information from past generations to predict behaviour(s) of the population in the current generation; observing the behaviours of each individual in the actual population, it rewards individuals that deviate from predicted behaviour(s): this is summarized in Fig.~\ref{fig:surpriseUML}. The following sections will describe the models of prediction and deviation, and the parameters which can affect their behaviour.

\subsection{Model of Prediction}\label{sec:surprisesearch_prediction}

As shown in Fig.~\ref{fig:surpriseUML}, the predictive model uses local information from previous generations to estimate (in a quantitative way) the expected behaviour(s) in the current population. 
Formally, predicted behaviours ($\mathbf{p}$) are created via eq.~\eqref{eq:prediction}, where $m$ is the predictive model that uses a degree of local (or global) behavioural information (expressed by $k$) from $h$ previous generations. Each parameter ($m$, $h$, $k$) may influence the scope and impact of predictions, and are problem-specific both from a theoretical (as they can affect performance of surprise search) and a practical (as certain domains may limit the possible choice of parameters) perspective.
\begin{equation}
\mathbf{p} = m(h, k) \label{eq:prediction}
\end{equation}
\subsubsection*{How much history of prior behaviours ($h$) should surprise search consider?}
In order to predict behaviours in the current population, the predictive model must consider previous generations. In order to estimate behaviours of a population at generation $t$, the predictive model must find trends in the populations of generations $t-1, t-2, \cdots, t-h$. The minimum number of generations to consider in order to observe an evolutionary trend, therefore, is $h=2$ (the two prior generations to the one being evaluated). However, behaviours that have performed well in the past could also be included in a \emph{surprise archive}, similar to the novelty archive of novelty search \cite{lehman2011abandoning}, and subsequently used to make predictions of interesting future behaviours. Such a surprise archive would serve as a more persistent history ($h>2$) but considering only the interesting historical behaviours rather than all past behaviours.

\subsubsection*{How local ($k$) are the behaviours surprise search needs to consider to make a prediction?}
Surprise search can consider behavioural trends of the entire population when creating a prediction (global information). In that case, $k=1$ and all behaviours are aggregated into a meaningful average metric for each prior generation. The current generation's expected behaviours are similarly expressed as a single (average) metric; deviation of individuals in the actual population is derived from that single metric. At the other extreme, surprise search can consider each individual in the population and derive an estimated behaviour based on the behaviours of its ancestors in the genotypic sense (parents, grandparents etc.) or behavioural sense (past individuals with the closest behaviour). In this case $k=P$ where $P$ is the size of the population, and the number of predictions to deviate from will similarly be $P$. Therefore, the parameter $k$ determines the level of \emph{prediction locality} which can vary from 1 to $P$; intermediate values of $k$ split prior populations into a number of population groups using problem-specific criteria and clustering methods.

\subsubsection*{What predictive model ($m$) should surprise search use?}
Any predictive modelling approach can be used to predict a future behaviour, such as a simple linear regression of a number of points in the behavioural space, non-linear extrapolations, or machine learned models. Again, we consider the predictive model, $m$, to be problem-dependent and contingent on the $h$ and $k$ parameters.

\subsection{Model of Deviation}\label{sec:surprisesearch_deviation}

To put pressure on unexpected behaviours, we need an estimate of the deviation of an observed behaviour from the expected behaviour (if $k=1$) or behaviours.
Following the principles of novelty search \cite{lehman2011abandoning}, this estimate is derived from the \emph{behaviour space} as the average distance to the $n$-nearest expected behaviours (prediction points). The \emph{surprise score} $s$ for an individual $i$ in the population is calculated as:
\begin{equation}
s(i) = \frac{1}{n}\sum_{j=0}^{n}d_s(i,p_{i,j})\label{eq:surprise}
\end{equation}
\noindent where $d_s$ is the domain-dependent measure of behavioural difference between an individual and its expected behaviour, $p_{i,j}$ is the $j$-closest prediction point (expected behaviour) to individual $i$ and $n$ is the number of prediction points considered; $n$ is a problem-dependent parameter determined empirically ($n{\leq}k$).

\subsection{Important notes}

\newadditions{
The evolutionary dynamics of surprise search are similar to those achieved via novelty search. A temporal window of where the search has been is maintained by looking at the prior behaviours (expressed by $h$ and $k$), which are used to make a prediction of the expected behaviours. However, surprise search works on a different search space (the prediction space) which is orthogonal to the behavioural space used by novelty, as it deviates from the expected and not from the actual behaviours. Therefore, a new form of divergent search emerges, which looks at previous behaviours only in an implicit way in order to derive the prediction space. A concern could be if surprise search is merely a version of random walk, especially considering that it deviates from predictions which can differ from actual behaviours. Several comparative experiments in section \ref{sec:experiments} show that surprise search is different and far more effective compared to various random benchmark algorithms.
}
The source code of the surprise search algorithm is publicly available here\footnote{http://www.autogamedesign.eu/mazesurprisesearch}.

\begin{figure}[t]
\centering
\subfloat[Neural Network]{
\includegraphics[width=0.45\linewidth]{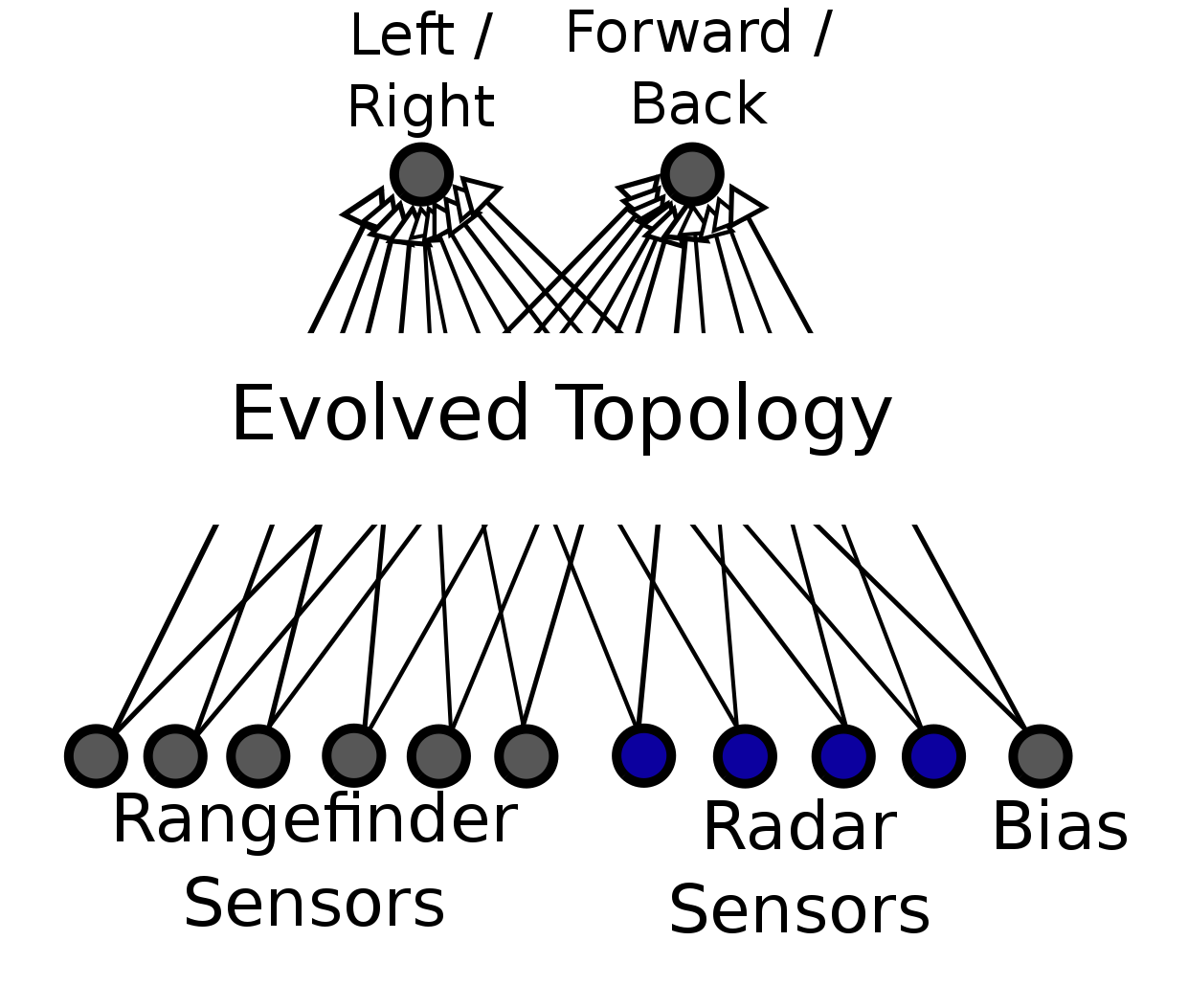}
\label{fig:ANN}}
\subfloat[Sensors]{
\includegraphics[width=0.43\linewidth]{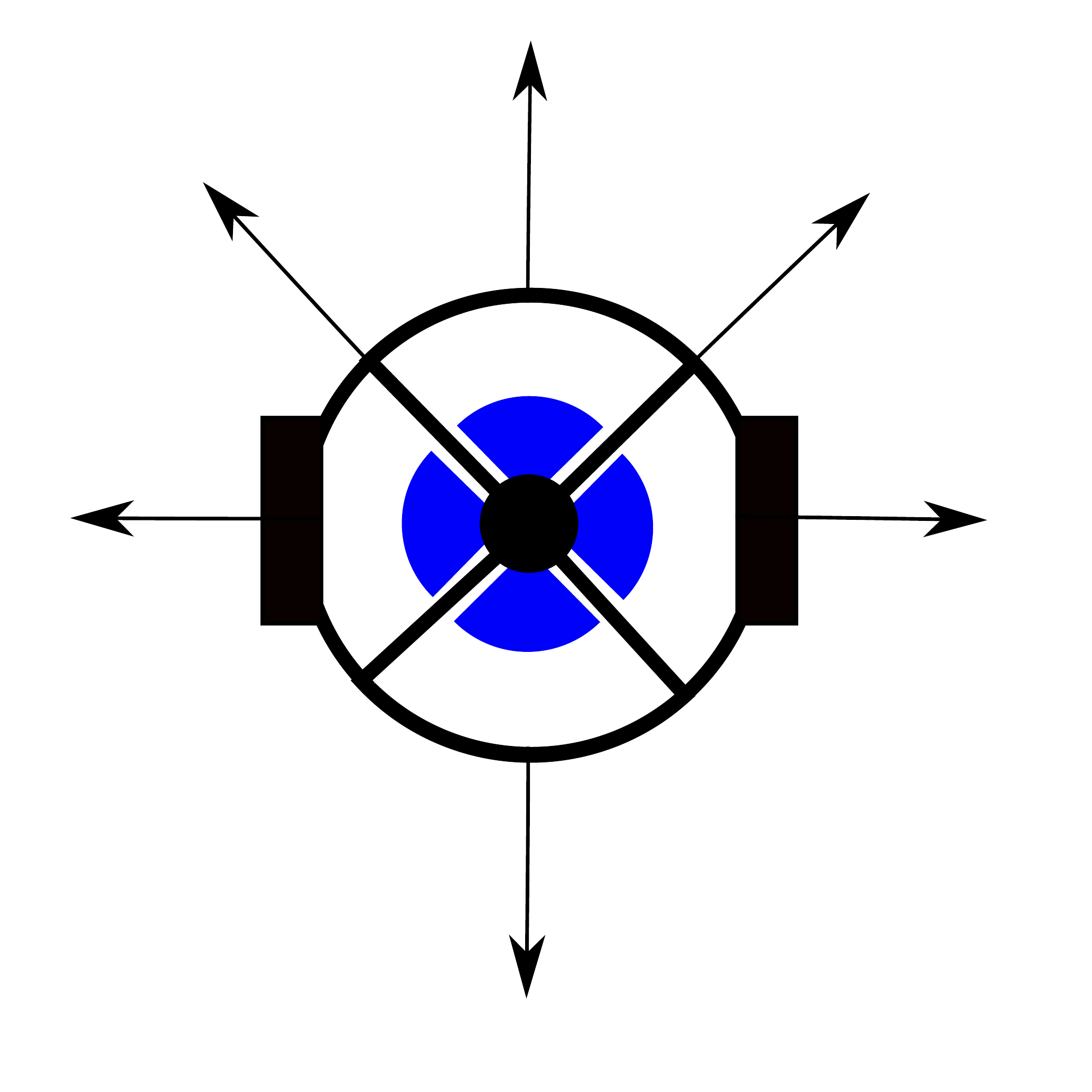}
\label{fig:robot}}
\caption{Robot controller for the maze navigation task. Fig.~\ref{fig:ANN} shows the network's inputs and outputs. Fig.~\ref{fig:robot} shows the layout of the sensors: the six black arrows are rangefinder sensors, and the four blue pie-slice sensors act as a compass towards the goal.}
\label{fig:NEAT}
\end{figure}

\section{Maze Navigation Test Bed}\label{sec:mazes}

\newadditions{
Inspired by the work of Lehman and Stanley for testing novelty search \cite{lehman2011abandoning}, we use a maze navigation problem as a testbed for surprise search, as it particularly suits the definition of deceptive problem.  The maze navigation task is made of a closed two-dimensional maze, which contains a start position and a goal position: the navigation task has a deceptive landscape---which is directly visible to a human observer---due to the several local optima present in the search space of the problem. \emph{Cul-de-sacs} added in the shortest path to the goal makes the problem more deceptive, as EC must visiting positions with lower fitness scores before reaching the goal, making the problem harder and more deceptive. In this case, navigation is performed by virtual robot controllers with sensors and mechanisms for controlling their direction and speed: the mapping between the two is provided is via an artificial neural network (ANN) evolved via neuroevolution of augmenting topologies \cite{stanley2002neat}.
Starting from the mazes introduced in \cite{lehman2011abandoning}, we designed two additional mazes of enhanced complexity and deceptiveness.
This section briefly describes the maze navigation problem, the four mazes adopted, and the parameters for the experiment of Section \ref{sec:experiments}.
}

\subsection{The Maze Navigation Task}\label{sec:mazes_task}

\newadditions{
The maze navigation task consists of finding the path from a starting point to a goal in a two-dimensional maze, in a fixed number of simulation steps. The problem becomes harder when mazes include dead-ends and the goal is far away from the starting point. 
As in \cite{lehman2011abandoning}, the robot has six range sensors to measure its distance from the closest obstacle, plus four range radars that fire if the goal is in their arc (see Fig. \ref{fig:robot}). Therefore, the robot's ANN receives 10 inputs from the sensors and it controls two actuators, i.e. whether to turn or change the speed (see Fig. \ref{fig:ANN}).
Evolving a controller able to successfully navigate a maze is a challenging problem, as EC needs to evolve a complex mapping between the input (sensors) and the output (movement) in an unknown environment. 
Even if it can be considered a toy problem, it is an interesting testbed as it stands for a general deceptive search space. Two properties have made this environment a canonical test for divergent search (e.g. \cite{lehman2011abandoning, pugh2016quality}): the ease of manually designing deceptive mazes and the low computational burden, which enables researchers to run multiple comparative tests among algorithms. Furthermore, the generality of the findings can be tested with automatically generated mazes, as in Section \ref{experiments_maze_generation}.
}

\subsection{Mazes}

\begin{figure}[t]
\centering
\subfloat[Medium]{
\includegraphics[width=70px]{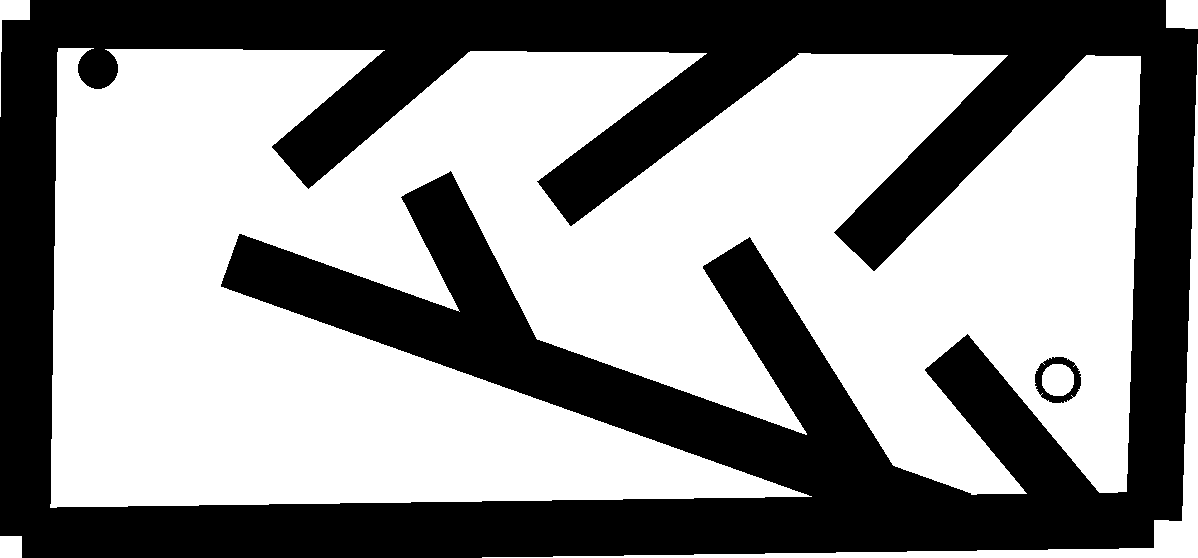}
\label{fig:medium}}
\subfloat[Hard]{
\includegraphics[width=45px]{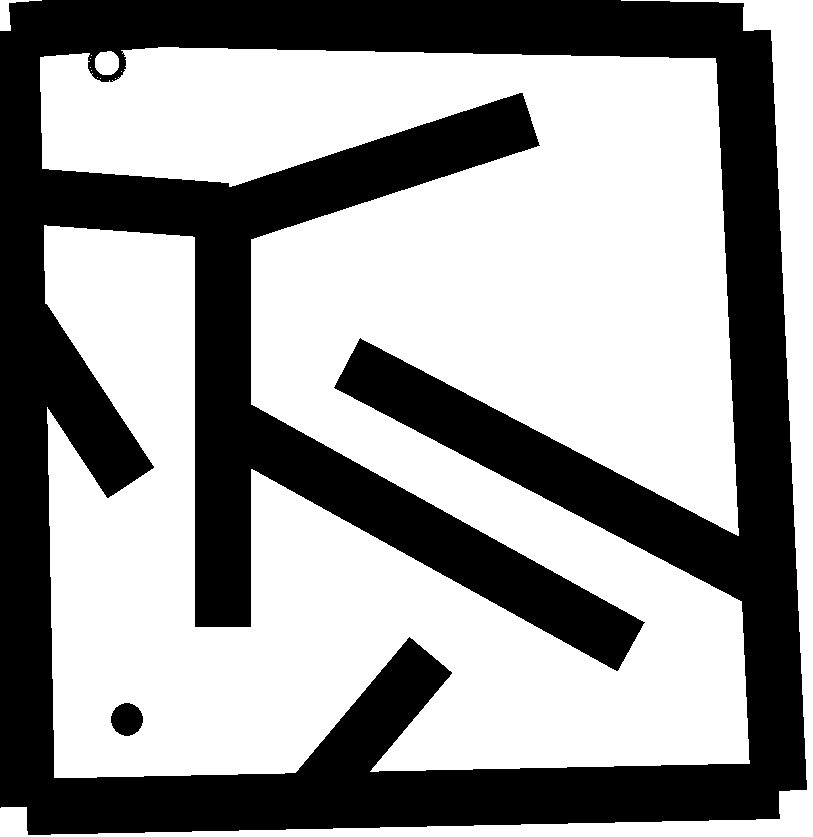}
\label{fig:hard}}
\subfloat[Very hard]{
\includegraphics[width=45px]{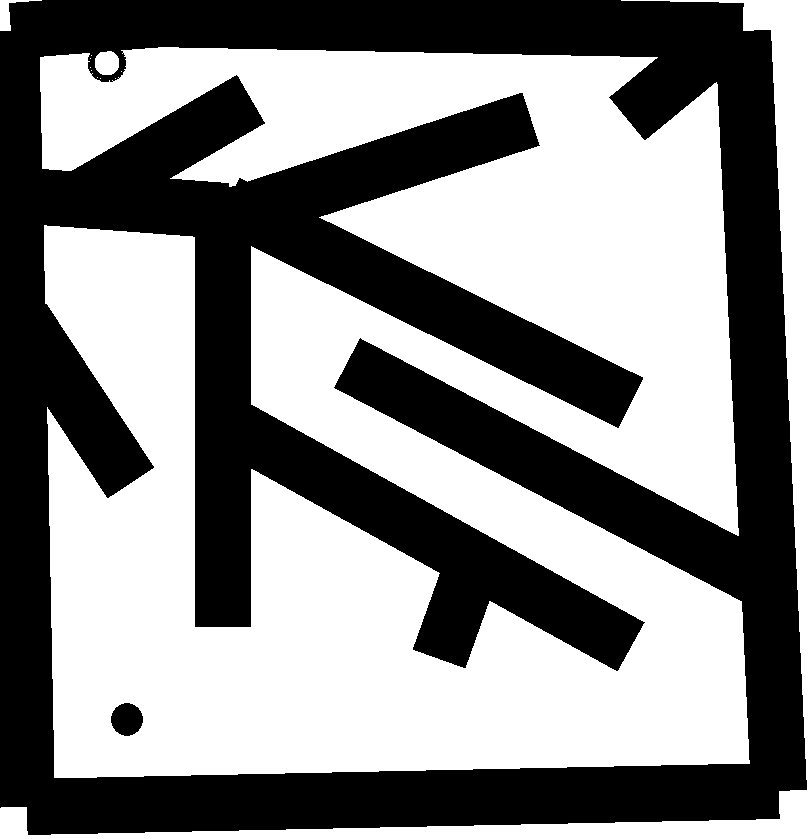}
\label{fig:very_hard}}
\subfloat[Extremely hard]{\makebox[60px]{
\includegraphics[width=45px]{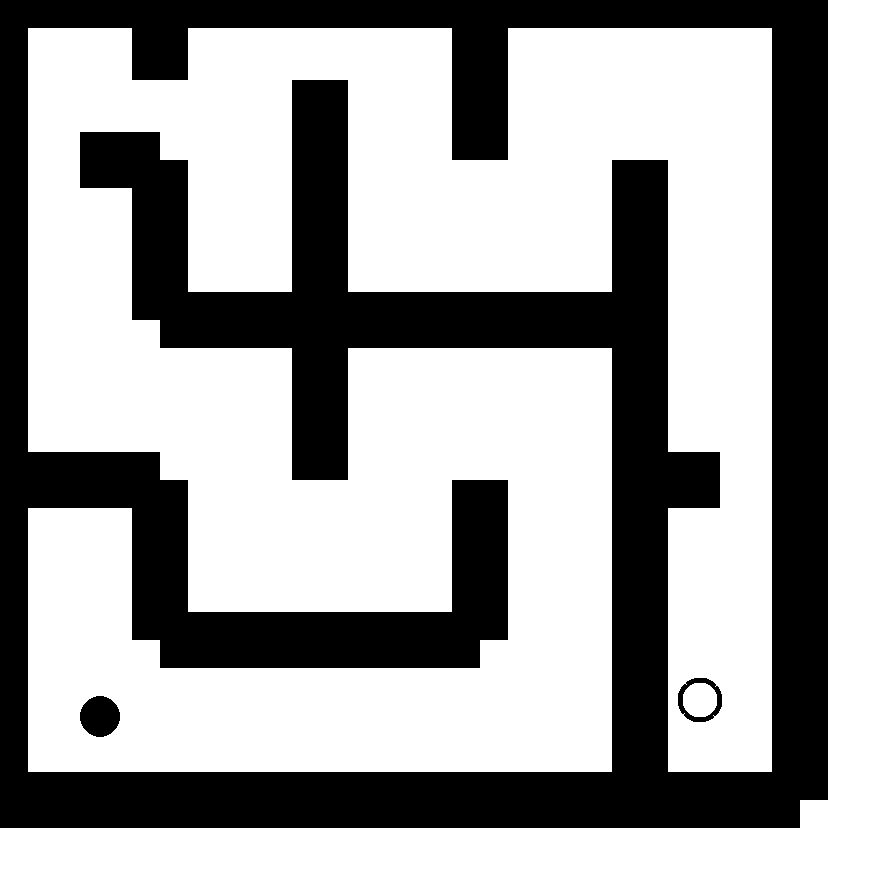}}
\label{fig:extreme_hard}}
\caption{The maze testbeds that appear in \cite{lehman2011abandoning} (Fig.~\ref{fig:medium} and \ref{fig:hard}) and new mazes  introduced in \cite{gravina2017noveltysurprise} and this paper (Fig.~\ref{fig:very_hard} and \ref{fig:extreme_hard} respectively). The filled circle is the robot\rq{}s starting position and the empty circle is the goal. The maze size is $300\times150$ units for medium and $200\times200$ for the other mazes. 
}
\label{fig:mazes}
\end{figure}

This paper tests the performance of surprise search on four mazes (see Fig.~\ref{fig:mazes}), two of which (\emph{medium} and \emph{hard}) have been used in \cite{lehman2011abandoning}. The medium maze (see Fig.~\ref{fig:medium}) is somewhat challenging as an algorithm should evolve a robot that avoids dead-ends placed alongside the path to the goal. The hard maze (see Fig.~\ref{fig:hard}) is more deceptive, due to the dead-end at the leftmost part of the maze; an algorithm must search in less promising (less fit) areas of the maze to find the global optimum. For these two mazes we follow the experimental parameters set in \cite{lehman2011abandoning} and consider a robot successful if it manages to reach the goal within a radius of five units at the end of an evaluation of 400 simulation steps. 

Beyond the two mazes of \cite{lehman2011abandoning}, two additional mazes (\emph{very hard} and \emph{extremely hard}) were designed to test an algorithm\rq{}s performance in even more deceptive environments. The very hard maze (see Fig. \ref{fig:very_hard}) is a modification of the hard maze introduced in \cite{gravina2017noveltysurprise} with more dead ends and winding passages. The extremely hard maze is a new maze (see Fig. \ref{fig:extreme_hard}) that features a longer and more complex path from start to goal, thereby increasing the deceptive nature of the problem. 

If we define a maze\rq{}s complexity as the shortest path between the start and the goal, complexity increases substantially from medium (240 units), to hard (360 units), to very hard (442 units) and finally to the extremely hard maze (552 units); note that the last three mazes are of equal size. The high problem complexity of the very hard and the extremely hard mazes led us to empirically increase the number of simulation steps for the evaluation of a robot to 500 and 1000 simulation steps, respectively. By increasing the simulation time in the more deceptive mazes we manage to achieve reasonable performances for at least one algorithm examined which allows for a better analysis and comparison.

\section{Algorithm Parameters for Maze Navigation}\label{sec:experiments_parameters}

This section provides details about the general and specific parameters for all the algorithms compared. We primarily test the performance of three algorithms: objective search, novelty search and surprise search, and include three baseline algorithms for comparative purposes. All algorithms use NEAT to evolve a robot controller with the same parameters as in \cite{lehman2011abandoning}, where the maze navigation task and the mazes of Fig.~\ref{fig:mazes} were introduced. Evolution is carried on a population of 250 individuals for a maximum of 300 generations in the medium and hard maze for a fair comparison to results obtained in \cite{lehman2011abandoning}. However, the number of generations is increased to 1000 for the more deceptive mazes (very hard and extremely hard) to allow us to analyse the algorithms\rq{} behaviour over a longer evolutionary period. 
The NEAT algorithm uses speciation and recombination, as described in \cite{stanley2002neat}. The specific parameters of all compared algorithms are detailed below. 

\subsection{Objective search}\label{sec:experiments_objective}

Objective search uses the agent's proximity to the goal as a measure of its fitness. Following \cite{lehman2011abandoning}, proximity is measured as the Euclidean distance between the goal and the position of the robot at the end of the simulation. This distance does not account for the maze's topology and walls, and can be deceptive in the presence of dead-ends.

\subsection{Novelty Search}\label{sec:experiments_novelty}

Novelty search uses the same novelty metric and parameter values as presented in \cite{lehman2011abandoning}. In particular, the novelty metric is the average distance of the robot from the nearest neighbouring robots among those in the current population and in a novelty archive. \emph{Distance} in this case is the Euclidean distance between two robot positions at the end of the simulation; this rewards robots ending in positions that no other robot has explored yet. The parameter for the novelty archive (e.g. the initial novelty threshold for inserting individuals to the archive is 6) is as given in \cite{lehman2011abandoning}.

\subsubsection*{Sensitivity Analysis}

While in \cite{lehman2011abandoning} novelty is calculated as the average distance from the $15$ nearest neighbours, the introduction of new mazes in this paper mandates that the $n$ parameter of novelty search is tested empirically. For that purpose we vary $n$ from $5$ to $30$ in increments of $5$ across all mazes and select the $n$ values that yield the highest number of maze solutions (successes) in 50 independent runs of 300 generations for the medium and hard maze, and 1000 generations for the other mazes. If there is more than one $n$ value that yields the highest number of successes then the lowest average evaluations to solve the maze is taken into account as a selection criterion. Figure \ref{fig:novelty_sensitivity} shows the results obtained by this analysis across all mazes. 

The best results are indeed obtained with $15$ nearest neighbours for the medium and hard maze, as in \cite{lehman2011abandoning} (49 and 48 successes, respectively). In the very hard maze there is no difference between $10$ and $15$ in terms of successes (39) but $n=15$ yields less evaluations, while in the extremely hard maze $n=10$ yields less evaluations and more successes (24) than any other value tested. In summary, reported results in Section \ref{sec:experiments} use $n=15$ for the medium, hard and very hard maze, and $n=10$ for the extremely hard maze.

\subsection{Surprise search}\label{sec:experiments_surprise}

Surprise search uses the surprise metric of eq.~\eqref{eq:surprise} to reward \emph{unexpected behaviours}. As with the other algorithms compared, \emph{behaviour} in the maze navigation domain is expressed as the position of the robot at the end of a simulation. The behavioural difference $d_s$ in eq.~\eqref{eq:surprise} is the Euclidean distance between the robots' final position and a considered prediction point, $p$. 

\begin{figure}[!tb]
\centering
\subfloat[Generation $t - 2$]{
\includegraphics[width=0.32\linewidth]{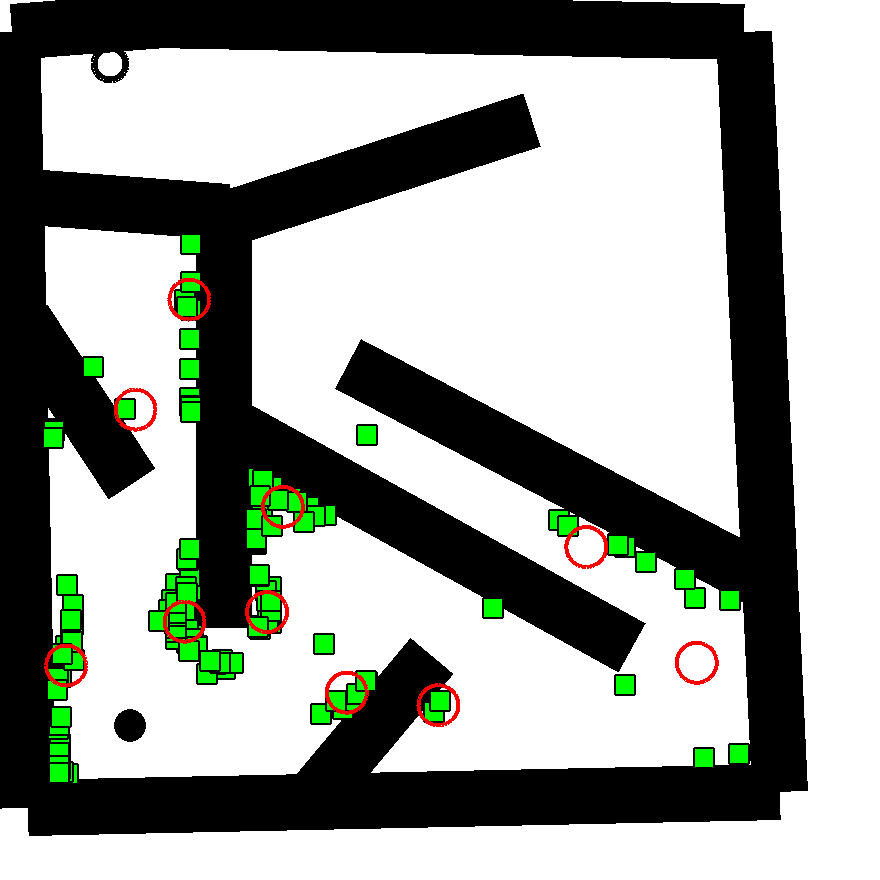}}
\subfloat[Generation $t - 1$]{
\includegraphics[width=0.32\linewidth]{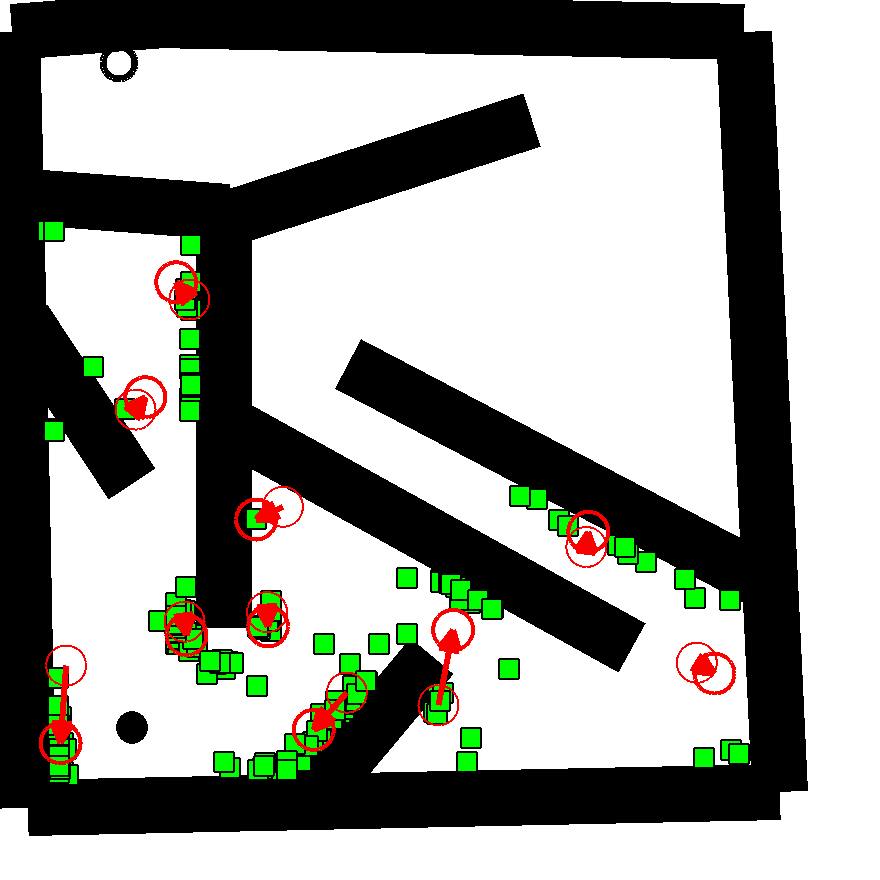}}
\subfloat[Generation $t$]{
\includegraphics[width=0.32\linewidth]{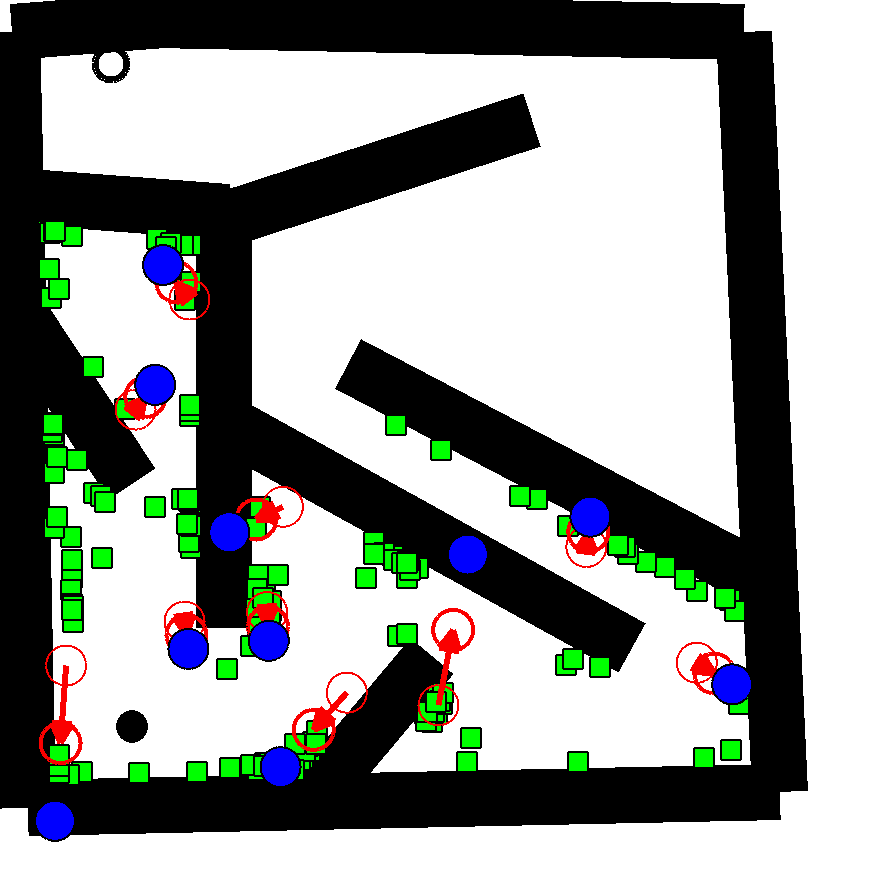}}
\caption{The key phases of the surprise search algorithm as applied to the maze navigation domain. Surprise search uses a history of two generations ($h=2$) and 10 behavioural clusters ($k=10$) in this example. Cluster centroids and prediction points are depicted as empty red (light gray in grayscale) and solid blue (dark gray in grayscale) circles, respectively.}
\label{fig:surprise_search}
\end{figure}

Following the general formulation of surprise in Section \ref{sec:surprisesearch_prediction}, the prediction points are a function of a model $m$ that considers $k$ local behaviours of $h$ prior generations. In this comparative study we use the simplest possible prediction model ($m$) which is a one-step linear regression of two points ($h=2$) in the behavioural space. Thus, only the two previous generations are considered when creating prediction points to deviate from in the current generation (see Fig.~\ref{fig:surprise_search}). In the first two generations the algorithm performs mere random search due to a lack of prediction points. 

The locality ($k$) of behaviours is determined by the number of behavioural clusters in the population that is obtained by running $k$-means on the final robot positions. 
The surprise search algorithm applies $k$-means clustering at each generation by seeding the initial configuration of the $k$ centroids with the centroids obtained in the previous generation; this seeding process is omitted only in the first generation due to the lack of earlier clusters. This way the algorithm is able to pair centroids in subsequent generations and track their behavioural history. Using the $k$ pairs of centroids of the last two generations, the algorithm creates $k$ prediction points for the current generation through a simple linear projection. Surprise search rewards the behaviour that obtains a surprising outcome given the temporal sequence of the final points of the robot across generations. Surprise search is thus orthogonal to objective and novelty search as it rewards robots that visit areas outside the predicted space(s), without any explicit knowledge of the final goal.


It should be noted that for high values of $k$ the $k$-means algorithm might end up not assigning any data point to a particular cluster; the chance of this happening increases with $k$ and the sparseness of data (in particular in datasets containing outliers) \cite{hartigan1979algorithm}. Further, the seeding initialization procedure we follow for $k$-means in this domain aims to behaviourally connect centroids across generations so as to enable us to predict and deviate from the expected behaviour in the next generation. The adopted initialization procedure (i.e.~inheriting from centroids of the previous generation) does not guarantee that all seeded centroids will be allocated a robot position during the assignment step of $k$-means, as robot positions (data points) might change drastically from one generation to the next. 
In the case of surprise search, when an empty cluster appears in the current generation (i.e. in positions where a cluster existed in a past generation but not currently), then its prediction is not updated (i.e.~moved) until a final robot position gets close to the empty cluster's centroid. Predicted centroids that have not been recently updated (due to empty clusters) are still considered when calculating the surprise score, and indirectly act as an archive of earlier predictions. However, this archive is not persistent as the number of `archived' prediction points can increase or decrease during the course of evolution, and depends on $k$.

\subsubsection*{Sensitivity Analysis}

To choose appropriate parameters for $k$ (information locality) and $n$ (number of prediction points) in the prediction and deviation models respectively, a sensitivity analysis is conducted for all mazes. We obtain $k$ empirically by varying its value between 20 and $P$ in increments of 20 for each maze. We also test all $k$ for $n=1$ and $n=2$ in this paper. As in the sensitivity analysis for novelty search we select the $k$ and $n$ values that yield the highest number of successes in 50 independent runs. 
If there is more than one $k$, $n$ combination that yields the highest number of successes we select the combination that solves the maze in the fewest average evaluations.

\begin{figure}[t]

\addtocounter{subfigure}{-1} 

\begin{minipage}{0.23\textwidth}
\centering
\subfloat{
\includegraphics[width=\linewidth]{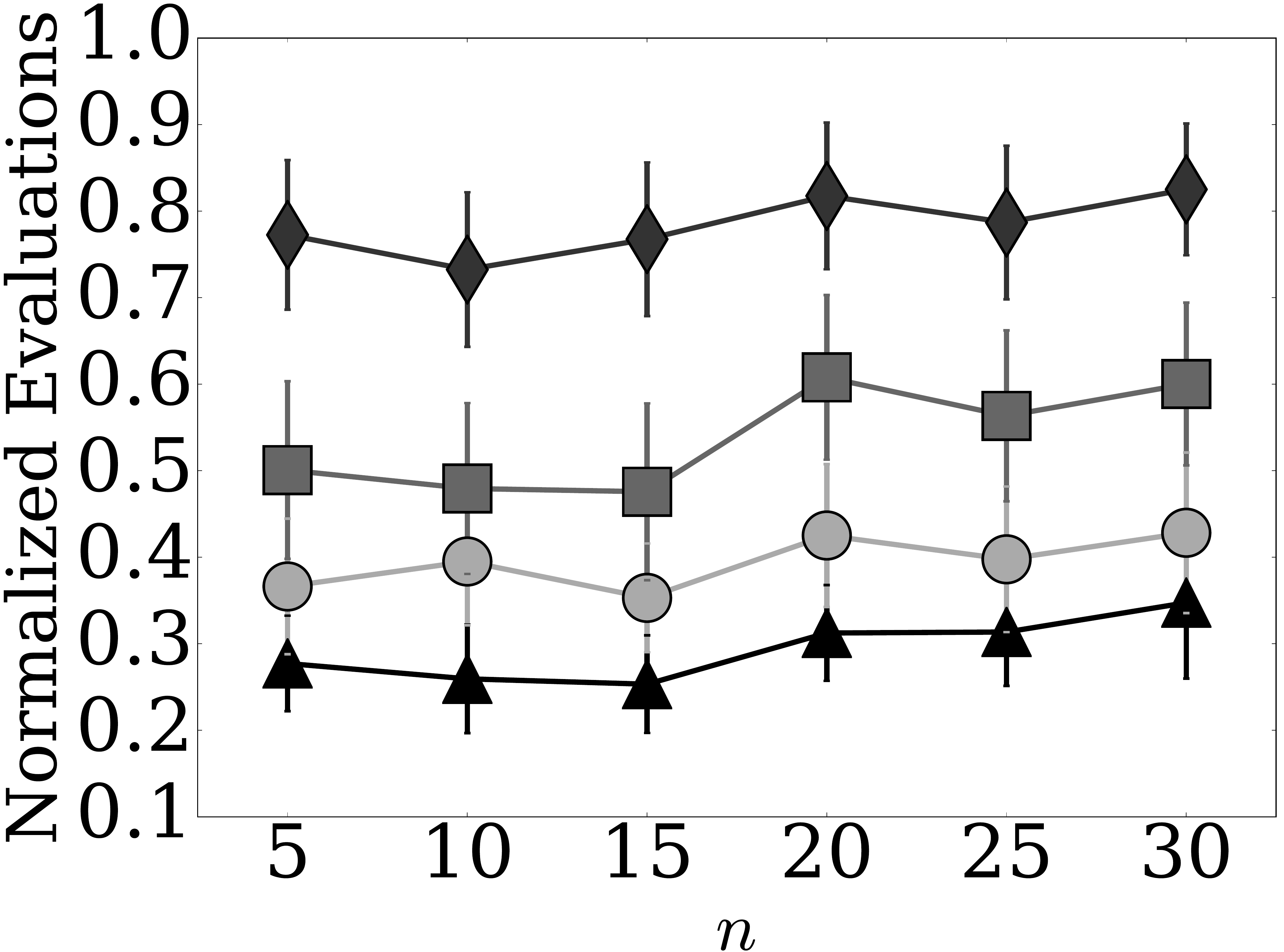}
}
\\
\subfloat[Novelty search parameters]{
\includegraphics[width=\linewidth]{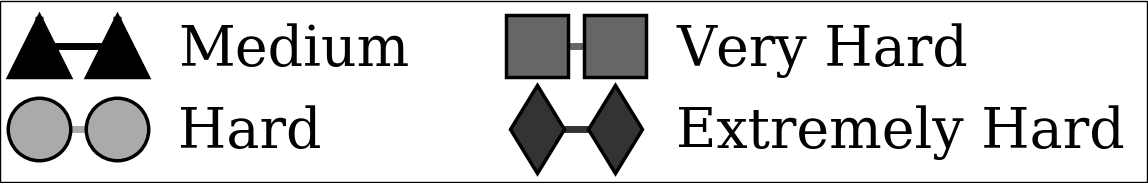}
\label{fig:novelty_sensitivity}
}
\end{minipage}~
\addtocounter{subfigure}{-1} 
\begin{minipage}{0.23\textwidth}
\centering
\subfloat{
\includegraphics[width=\linewidth]{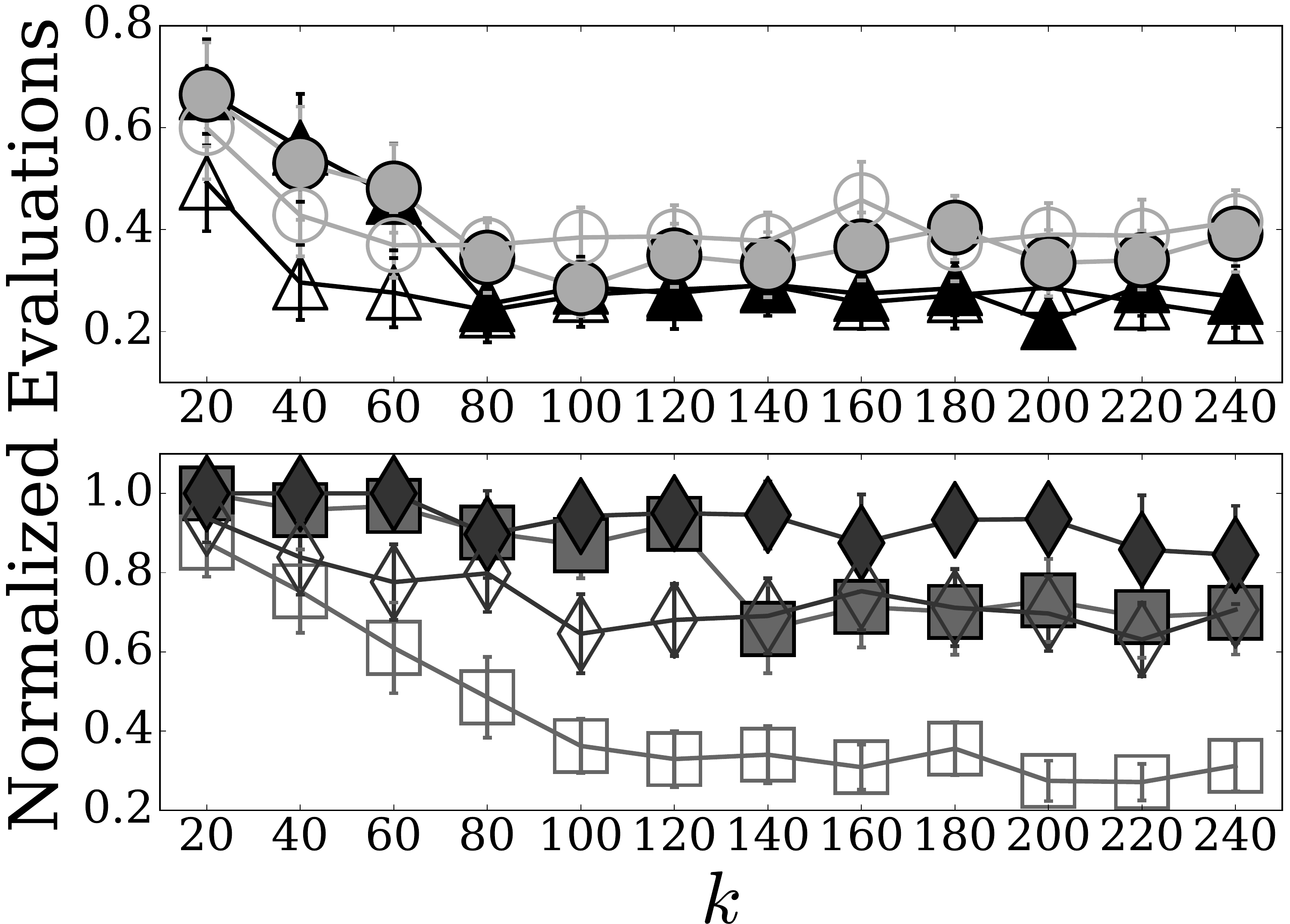}}
\\
\subfloat[Surprise search parameters]{
\includegraphics[width=\linewidth]{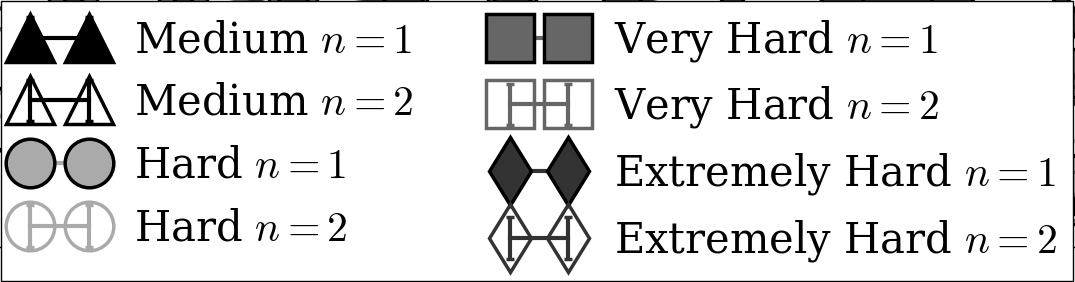}
\label{fig:surprise_sensitivity}
}
\end{minipage}

\caption{\textbf{Sensitivity Analysis:} selecting $n$ for novelty search (Fig.~\ref{fig:surprise_sensitivity}), $k$ and $n$ for surprise search (Fig.~\ref{fig:novelty_sensitivity}). Figures depict the average number of evaluations (normalized by the total number of evaluations) obtained out of 50 runs . Error bars represent the 95\% confidence interval.} 
\end{figure}

Figure \ref{fig:surprise_sensitivity} shows the average number of evaluations for all $k$ values tested, for $n=1$ and $n=2$. It is clear that higher $k$ values result to less evaluations on average. Moreover, it seems that $n = 2$ leads to better performance in the two more deceptive mazes. Based on the above selection criteria, we pick $k = 200$ and $n=1$ for the medium maze, which gives the highest number of successes (50) and the lowest number of evaluations ($16,364$ evaluations on average). For the hard maze we select $k=100$ and $n=1$, as it is the most robust (49 success) and fastest ($23,214$ evaluations on average) among tested values. Following the same procedure $k=200$ and $n=2$ in the very hard maze, and $k=220$ and $n=2$ in the extremely hard maze (see Fig. \ref{fig:surprise_sensitivity}).

In this paper we started our investigations with the simplest possible prediction model ($m$), which is a linear regression, and the shortest possible time window of two generations for the history parameter ($h$). The impact of the history parameter and the prediction model on the algorithm\rq{}s performance is not examined empirically in this paper and remains open to future investigations. We get back to this discussion in Section \ref{sec:discussion}. 

\subsection{Other baseline algorithms}\label{sec:experiments_baselines}

Three more baseline algorithms are included for comparative purposes. \emph{Random search} is a baseline proposed in \cite{lehman2011abandoning} that uses a uniformly-distributed random value as the fitness function of an individual. 
The other two baselines are variants of surprise search that test the impact of the predictive model. \emph{Surprise search (random)}, SS$_{r}$, selects $k$ random prediction points ($p_{i,j}$ in eq.~\ref{eq:surprise}) within the maze following a uniform distribution, and tests how surprise search would perform with a highly inaccurate predictive model. \emph{Surprise search (no prediction)}, SS$_{np}$, uses the current generation's actual clusters as its prediction points ($p_{i,j}$ in eq.~\ref{eq:surprise}), thereby, omitting the prediction phase of the surprise search algorithm. SS$_{np}$ uses real data (cluster centroids) from the current generation rather than predicted data regarding the current generation, and tests how the algorithm performs divergent search from real data. Note that SS$_{np}$ is reminiscent of novelty search, except that it uses deviation from cluster centroids (not points) and does not use a novelty archive. The same parameter values ($k$ and $n$) are used for these variants of surprise search.

\section{Surprise Search in Authored Deceptive Mazes}\label{sec:experiments}

The robot maze navigation problem is used to compare the performance of surprise, novelty and objective search. To test the algorithms' performance, we follow the approach proposed in \cite{yannakakis2003performance} and compare their \emph{efficiency} and \emph{robustness} in all four test bed mazes. We finally analyse some typical examples on both the behavioural and the genotypical space of the generated solutions.
All results reported are obtained from 100 independent evolutionary runs; reported significance and corresponding $p$ values are obtained via two-tailed Mann-Whitney U-test, with a significance level of 5\%. 

\subsection{Efficiency}\label{experiments_efficiency} 

\begin{figure}[!t]
\centering
\subfloat[Medium maze]{
\includegraphics[width=0.47\linewidth]{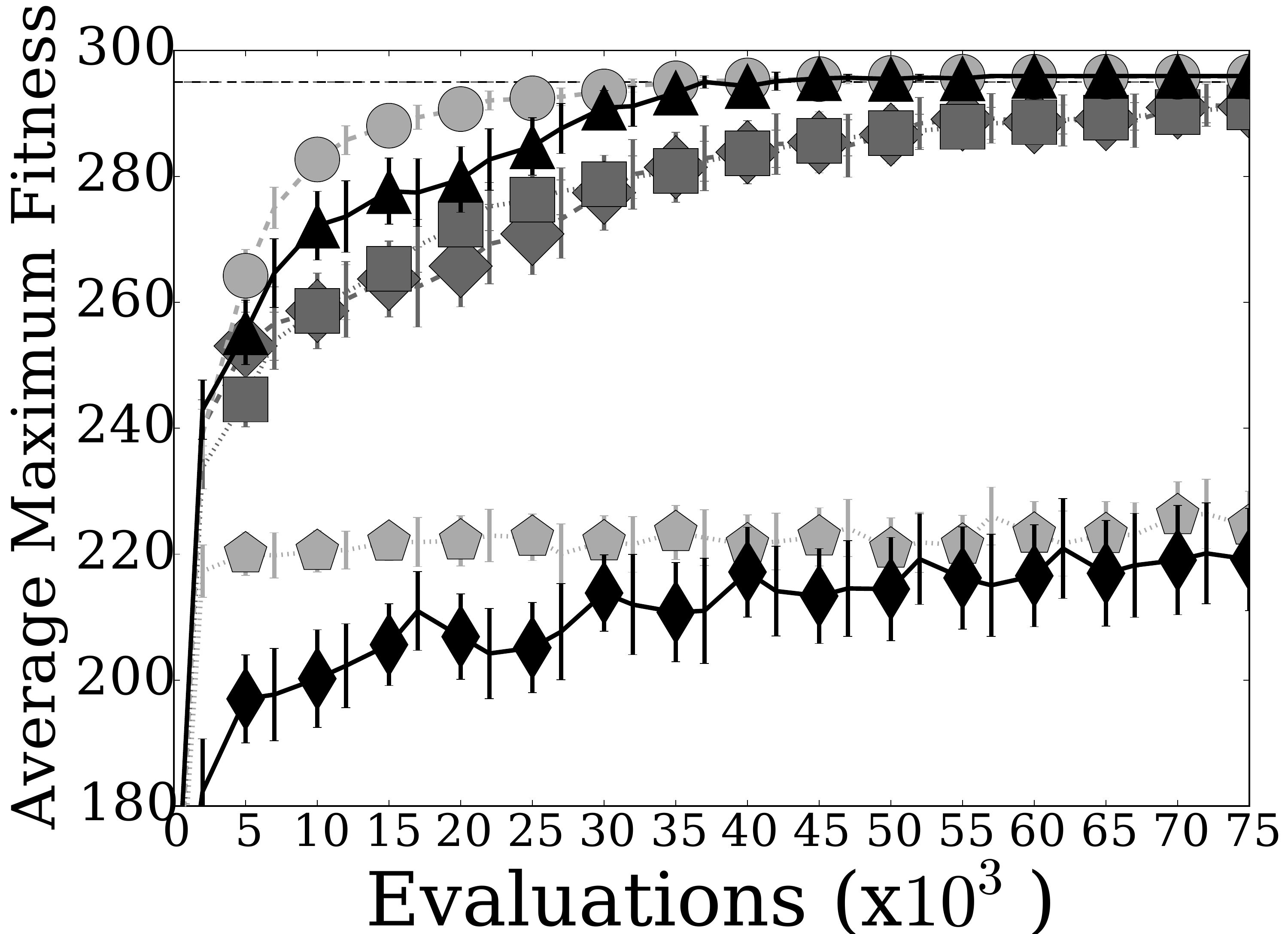}
\label{fig:fit_medium}}
\hfill
\subfloat[Hard maze]{
\includegraphics[width=0.47\linewidth]{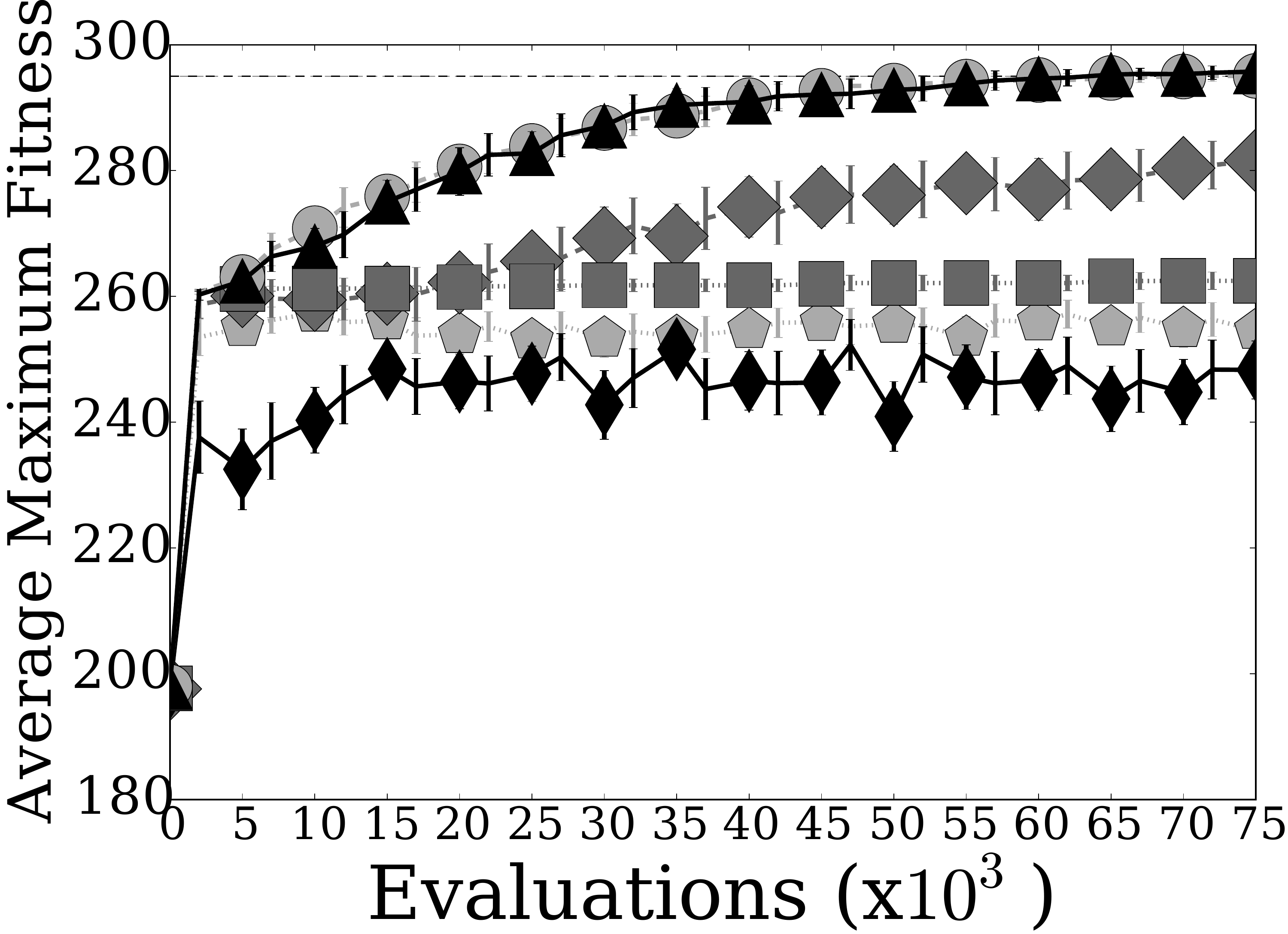}
\label{fig:fit_hard}}
\vfill
\subfloat[Very hard maze]{
\includegraphics[width=0.47\linewidth]{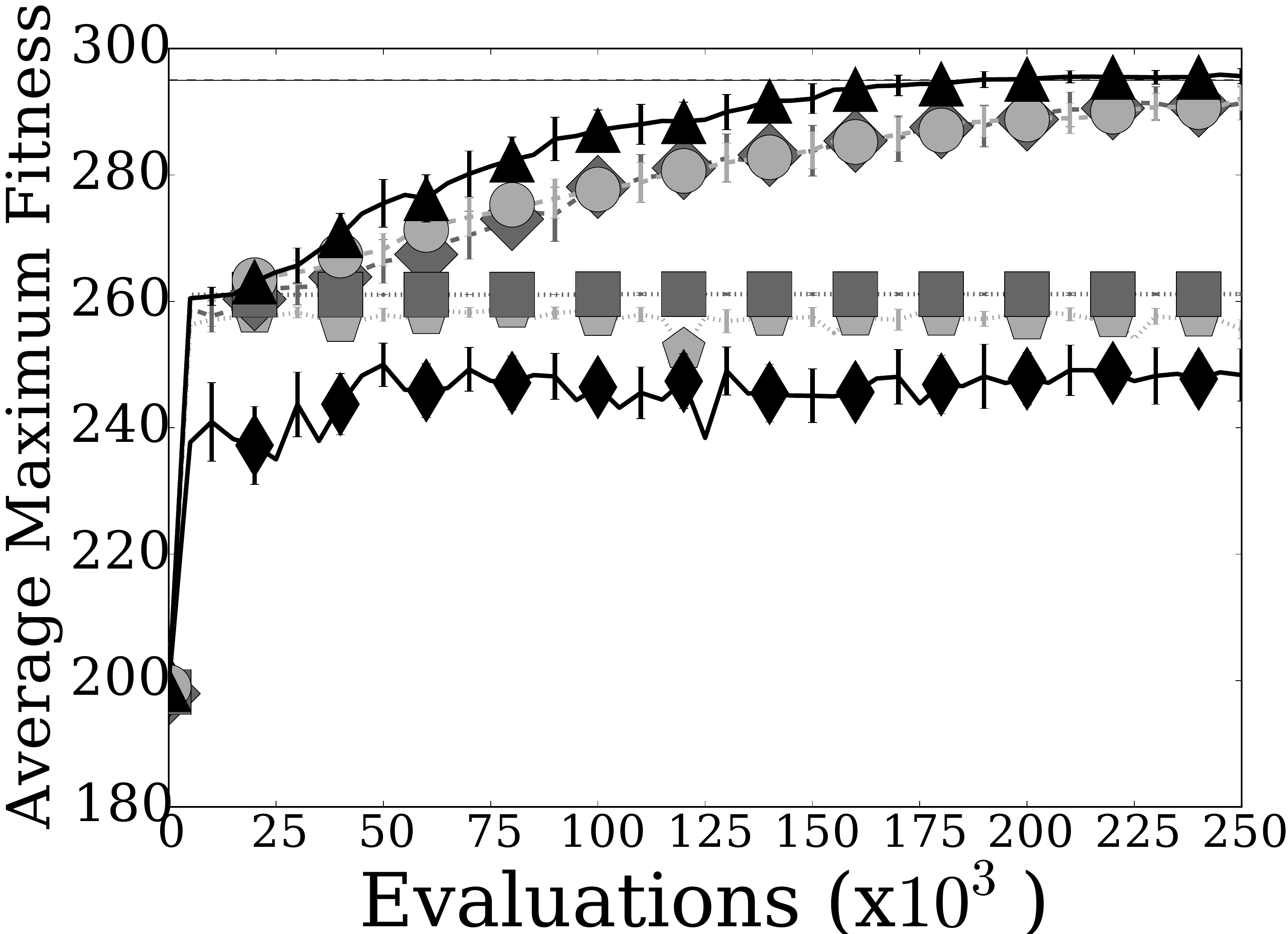}
\label{fig:fit_very_hard}}
\hfill
\subfloat[Extremely hard maze]{
\includegraphics[width=0.47\linewidth]{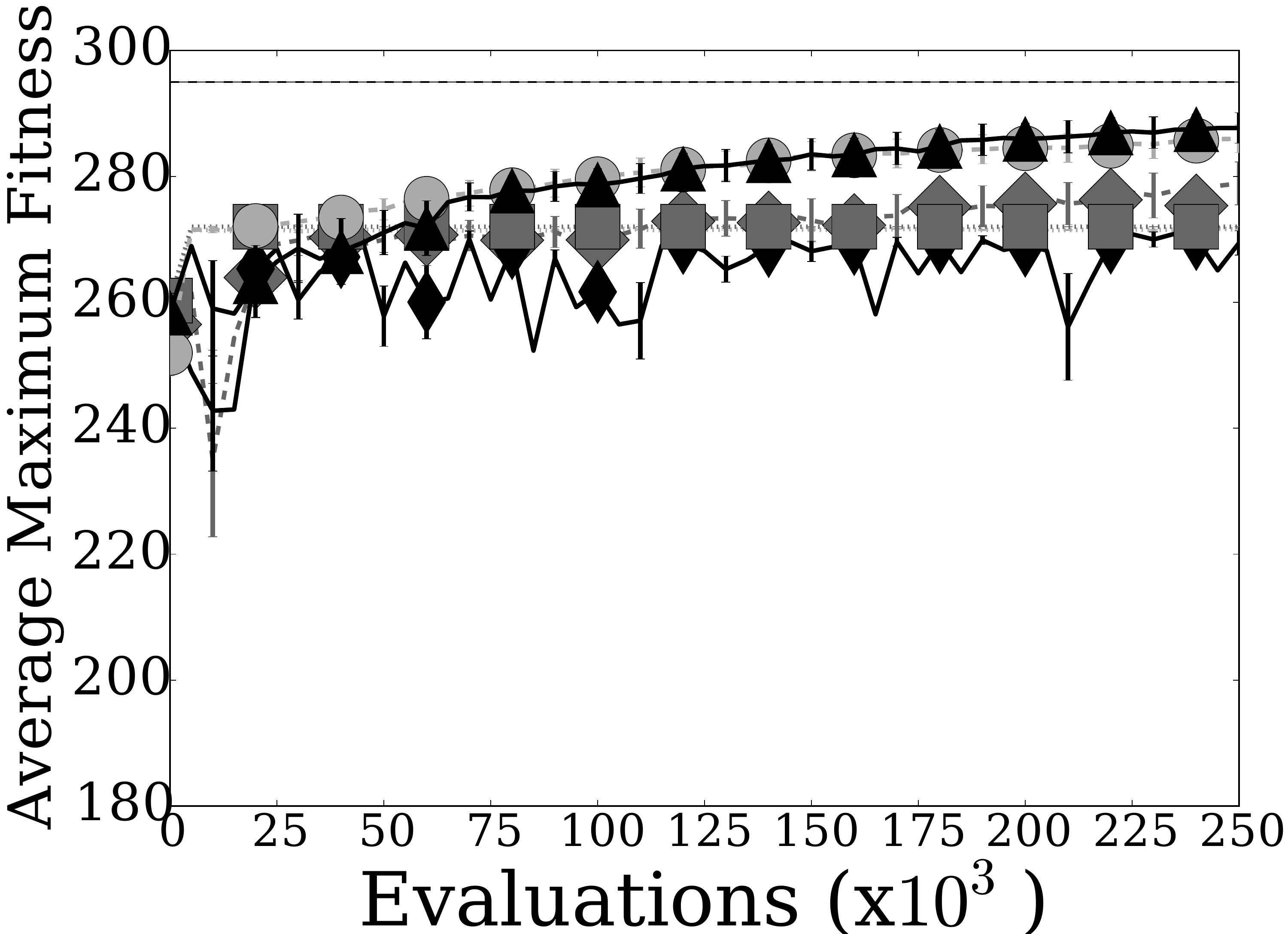}
\label{fig:fit_extreme_hard}} \\
\subfloat{
\includegraphics[width=0.85\linewidth]{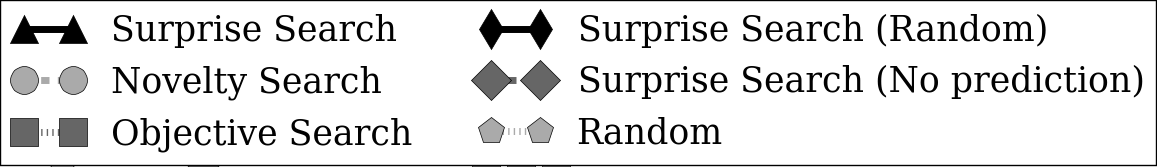}}
\caption{\textbf{Efficiency} (average maximum fitness) comparison for the four mazes in Fig.~\ref{fig:mazes}. The graphs depict the evolution of fitness over the number of evaluations. Values are averaged across 100 runs of each algorithm and the error bars represent the 95\% confidence interval of the average. \label{fig:fitness}}
\end{figure}

\newadditions{
Efficiency is defined as the \emph{maximum fitness over time}, where fitness is calculated as $300-d(i)$; $d(i)$ the Euclidean distance between the final position of robot $i$ and the goal, as in \cite{lehman2011abandoning}.
Figure \ref{fig:fitness} shows the average maximum fitness across evaluations for each approach for the four mazes. 
}

\newadditions{
In the medium maze, we can observe that both surprise and novelty search converge after approximately \num[group-separator={,}]{35000} evaluations. Even if novelty seems to yield a higher average maximum fitness values than surprise search, the difference is insignificant. Novelty search, on average, obtains a final maximum fitness of $295.84$ ($\sigma =  1.47$), while surprise search obtains a fitness of $295.93$ ($\sigma = 0.72$); $p > 0.05$. 
By looking at the 95\% confidence intervals, it seems that novelty search yields higher average maximum fitness between \num[group-separator={,}]{7500} and \num[group-separator={,}]{25000} evaluations. 
This difference is due to the predictions that surprise search tries to deviate from. Early during evolution, two consecutive generations may have robots far from each other, lead to distant and erratic predictions.
Eventually, we have a convergence and the predictions become more consistent, allowing surprise search to solve the maze.
Both objective search and SS$_{np}$ seem fairly efficient to solve the maze; however, they are not able to find the goal in all the runs. The random baselines, instead, perform poorly and show very little improvement as evolution progresses. The baselines' performance proves that surprise search is different from a random walk and that the prediction model positively affects the performance of the algorithm.
}

\newadditions{
In a more deceptive test, the hard maze, we can see from Fig.~\ref{fig:fit_hard} that novelty and surprise perform much better than all other algorithms; differences in efficiency between novelty and surprise search are not significant.
Surprise search and novelty search find the goal in 99 and 93 out of 100 runs respectively, SS$_{np}$ finds the solution in 61 runs, while for the rest of the baselines the success rate is far below.
It is interesting to note that objective search reaches a high fitness score around $260$ at the very beginning of the evolutionary process, and then it stops to improve. 
This is due to the dead-end at the upper right corner of the maze (Fig.~\ref{fig:hard}), which prevents the algorithm from discovering the global optimum. In order to discover the global optimum, in fact, the algorithm needs to explore the least fit areas of the search space, such as the bottom-right corner of the maze.
}

In the very hard maze, objective search never finds the solution in 100 runs; Fig.~\ref{fig:fit_very_hard} shows that this algorithm is not able to reach the goal because, as in the hard maze, it reaches the left-most dead end and is unable to bypass that local optimum. Unsurprisingly, the random and $SS_{R}$ baselines also perform poorly. On the other hand, novelty search finds the solution in 85 out of 100 runs, while surprise search finds a solution in 99 runs. Interestingly, SS$_{np}$ finds 88 solutions out of 100 runs, similar to novelty search. This can be explained by looking at how SS$_{np}$ is implemented. Its behaviour is quite similar to novelty search as it merely uses the local behaviours of the current generation; the key difference is that these behaviours are clustered in SS$_{np}$. 
In such a complex maze surprise search seems to handle maze deceptiveness in a better way; it obtains a final maximum fitness of $295.77$ ($\sigma = 3.59$) which is higher (but not significantly) than that of novelty search ($292.15$; $\sigma = 9.85$); $p > 0.05$. From the confidence intervals of Fig.~\ref{fig:rob_very_hard}, it appears that surprise search is performing better or significantly better than novelty search after $50,000$ evaluations until the end of the run. 

In the most deceptive (extremely hard) maze, objective search, random and SS$_{r}$ do not find any solution in 100 runs, performing poorly in terms of efficiency. This is not surprising as all of these algorithms perform consistently poorly in all but the simplest mazes. Novelty search and surprise search find the solution 48 and 67 times, respectively, while the SS$_{np}$ obtains 39 solutions. As can be seen in Fig.~\ref{fig:fit_extreme_hard}, surprise search yields a higher maximum fitness after \num[group-separator={,}]{150000} evaluations, with a final maximum fitness of $287.68$ ($\sigma = 12.16$).

\newadditions{
Another way of estimating efficiency is the effort it takes an algorithm to find a solution. In this case, surprise clearly manages to be more advantageous.
}
In the medium maze surprise search manages to find the goal, on average, in $16,084$ evaluations ($\sigma = 11,588$) which is faster than novelty ($19,814$; $\sigma = 15,441$) and significantly faster than objective search ($48,186$; $\sigma = 23,590$) and SS$_{np}$ ($26,452$; $\sigma = 21,249$). We observe the same comparative advantage in the hard maze as surprise search solves the problem in $23,566$ evaluations on average ($\sigma = 15,925$) whereas novelty search, SS$_{np}$, and objective search solve it in $28,493$ ($\sigma = 19,939$), $47,550$ ($\sigma = 25,524$) and $73,643$ ($\sigma = 7,542$) evaluations, respectively. Most importantly surprise search is significantly faster ($p < 0.05$) than novelty search in the more deceptive problems: on average surprise search finds the solution in $76,261$ evaluations ($\sigma = 52,385$) in the very hard maze and in $154,794$ evaluations ($\sigma = 84,733$) in the extremely hard maze, whereas novelty search requires $115,600$ evaluations ($\sigma = 81,091$) and $178,045$ evaluations ($\sigma = 86,410$), respectively. Furthermore surprise search is significantly faster ($p < 0.01$) than SS$_{np}$, which requires $117,560$  ($\sigma = 74,430$) and $200,190$ ($\sigma = 75,569$) evaluations in the very hard and the extremely hard maze, respectively. 

The findings from the above experiments indicate that, in terms of maximum fitness obtained, surprise search is comparable to novelty search and far more efficient than objective search in deceptive domains. We can further argue that the deviation from the predictions (which are neither random nor omitted) is beneficial for surprise search as indicated by the performances of SS$_{r}$ and SS$_{np}$. The performance of this baseline appears to be similar to novelty search, especially in harder mazes; this is not surprising as SS$_{np}$ is conceptually similar to novelty search, as noted in Section \ref{sec:experiments_baselines}. 
It is also clear that, on average, surprise search finds the solution faster than any other algorithm in all mazes.

\subsection{Robustness}\label{experiments_robustness} 

\begin{figure}[t]
\centering
\subfloat[Medium maze]{
\includegraphics[width=0.47\linewidth]{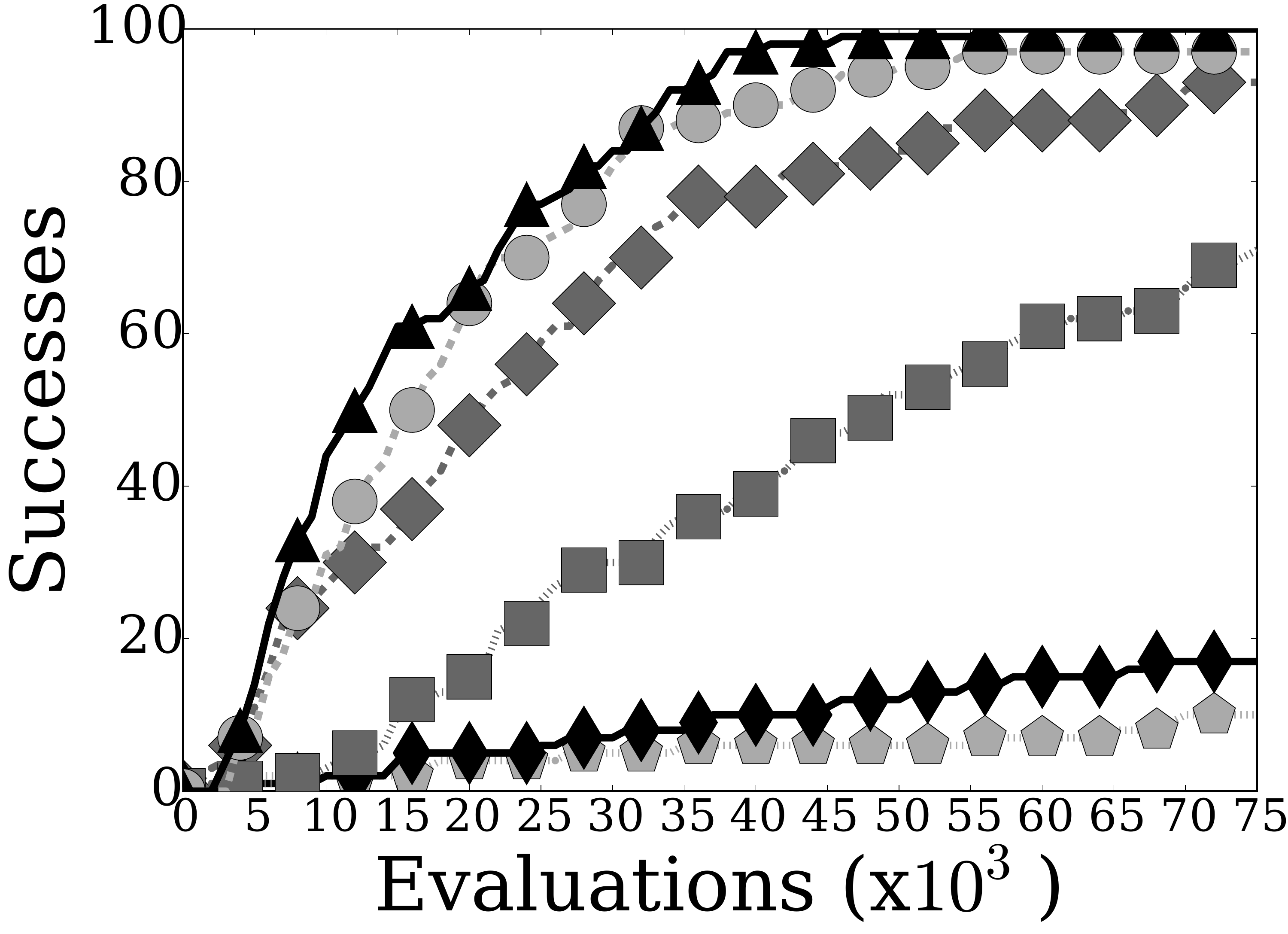}
\label{fig:rob_medium}}
\hfill
\subfloat[Hard maze]{
\includegraphics[width=0.47\linewidth]{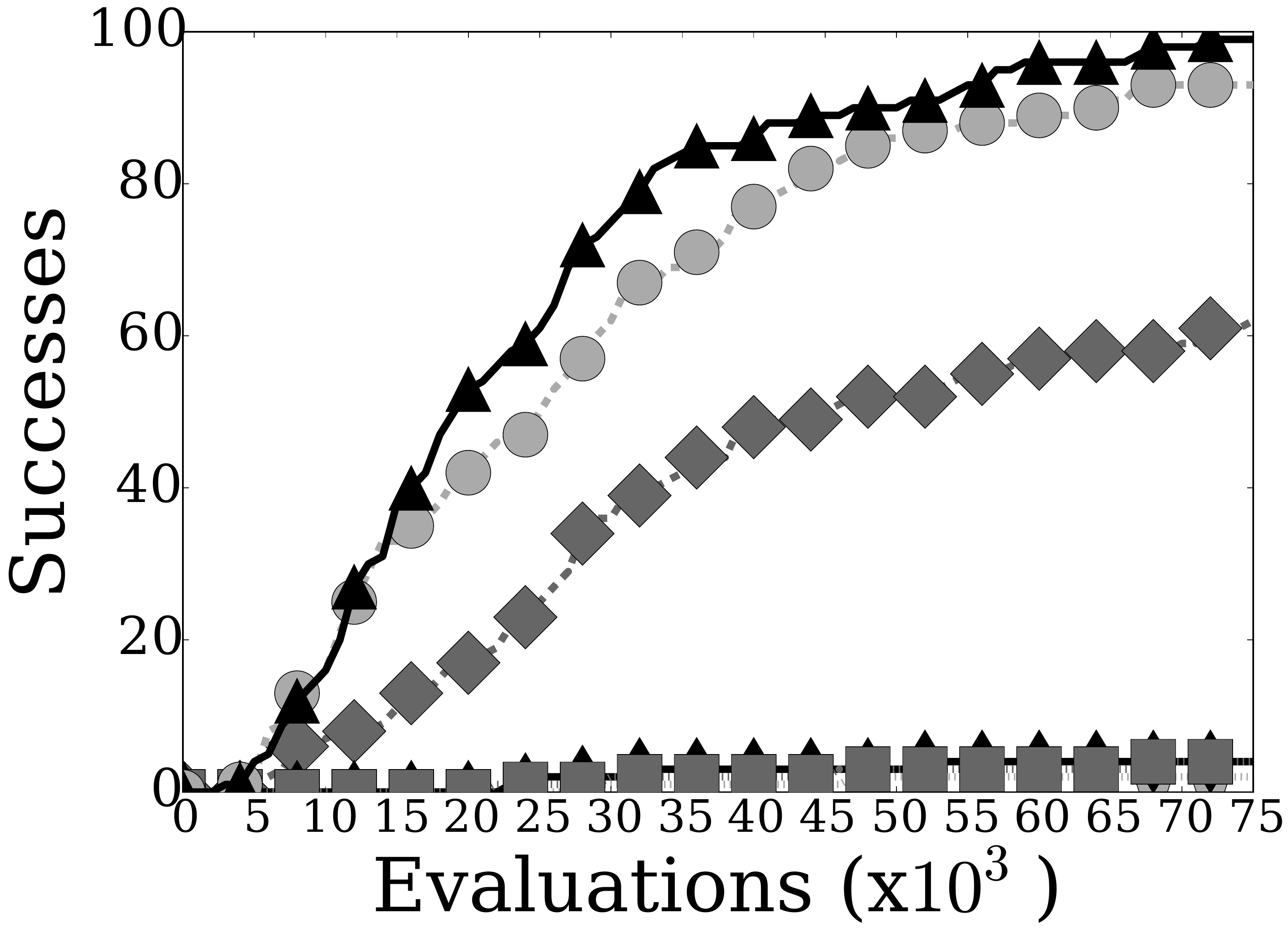}
\label{fig:rob_hard}}
\vfill
\subfloat[Very hard maze]{
\includegraphics[width=0.47\linewidth]{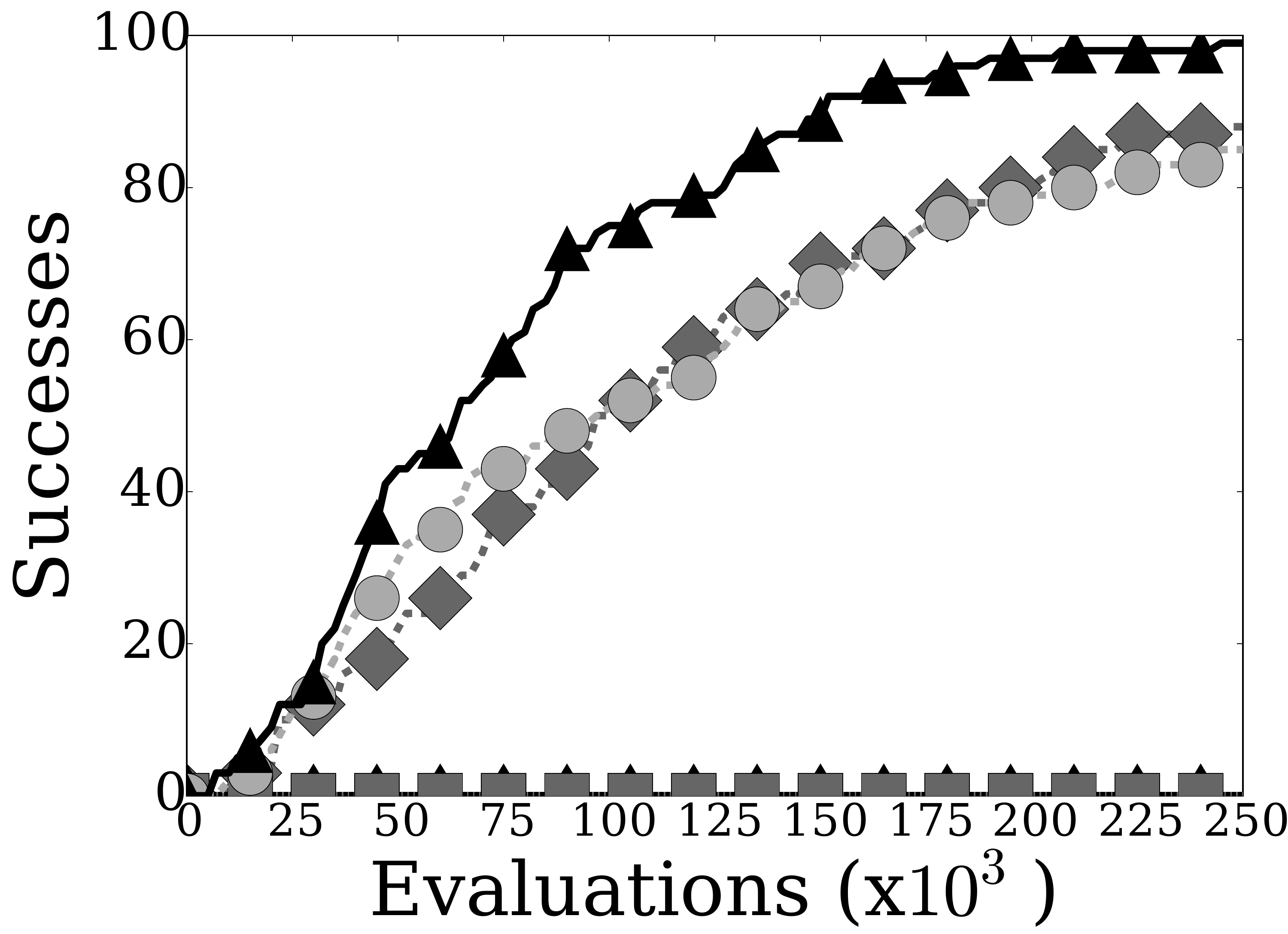}
\label{fig:rob_very_hard}}
\hfill
\subfloat[Extremely hard maze]{
\includegraphics[width=0.47\linewidth]{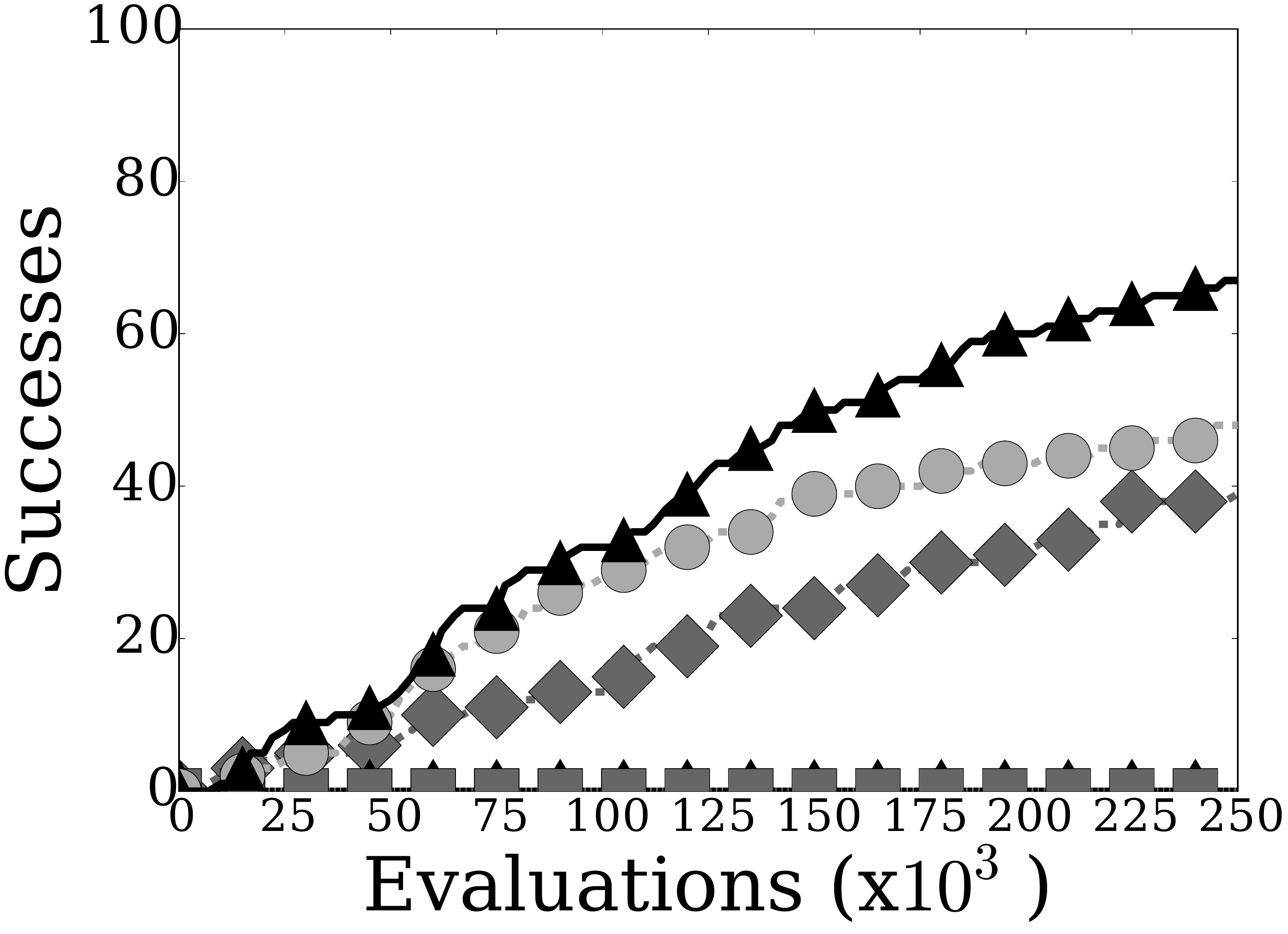}
\label{fig:rob_extreme_hard}} \\
\subfloat{
\includegraphics[width=0.85\linewidth]{./graph/legend_v2}}
\caption{\textbf{Robustness} comparison for the four mazes in Fig.~\ref{fig:mazes}. The graphs depict the evolution of algorithm successes in solving the maze problem over the number of evaluations. \label{fig:rob}}
\end{figure}

\newadditions{
\emph{Robustness} is defined as the number of successes obtained by the algorithm across time (i.e. evaluations). In Figure \ref{fig:rob} we compare the robustness of each approach across the four mazes, collected from 100 runs. In the medium maze (Fig. \ref{fig:rob_medium}), surprise search is more successful than novelty search in the first \num[group-separator={,}]{20000} evaluations; moreover, surprise search finds, on average, the 100 solutions in fewer evaluations compared to the other approaches. As noticed in the previous section, in the first \num[group-separator={,}]{20000} evaluations novelty search has a comparable or higher efficiency in Fig.~\ref{fig:fit_medium}; this points to the fact that while some individuals in surprise search manage to reach the goal, others do not get as close to it as in novelty search.
On the other hand, objective search fails to find the goal in 29 runs, because of the several dead-ends present in this maze. The control algorithm SS$_{np}$ finds the goal 93 times out of 100, but it's slower compared to novelty and surprise search. Few solutions are found by the baseline random search and SS$_{r}$, and they are significantly slower than the other approaches.
Fig.~\ref{fig:rob_hard} shows that, in the hard maze, novelty search attains more successes than surprise search in the first \num[group-separator={,}]{10000} evaluations but the opposite is true for the remainder of the evolutionary progress. As in the previous maze, this behaviour is not reflected in the efficiency graph (Fig.~\ref{fig:fit_hard}): this can be explained by how surprise search evolves individuals, as they change their distance to the goal more abruptly, while novelty search evolves behaviours in smooth incremental steps. On the other hand, SS$_{np}$ finds fewer solutions in this maze, 62 out of 100. Finally, the deceptive properties of this maze are exemplified by the poor performance of objective search and the two random baselines.
}

\begin{table*}[t]
\centering
\caption{\textbf{Behavioural Space.} Typical successful runs solved after a number of evaluations ($E$) across the four mazes examined. Heatmaps illustrate the aggregated numbers of final robot positions across all evaluations. Note that white space in the maze indicates that no robot visited that position. The entropy ($H \in [0,1]$) of visited positions is also reported and is calculated as follows: $H = (1/logC) \sum_i\{(v_i/V)log(v_i/V)\}$; where $v_i$ is the number of robot visits in a position $i$, $V$ is the total number of visits and $C$ is the total number of discretized positions (cells) considered in the maze.
\label{fig:heatmaps}}
\bgroup
\def\arraystretch{1}
    \begin{tabular}{|c  c | c c | c c | c c |}
 
   \hline
    \multicolumn{2}{|c}{\textbf{Medium Maze}} & 
    \multicolumn{2}{|c}{\textbf{Hard Maze}} & 
    \multicolumn{2}{|c}{\textbf{Very Hard Maze}} &
    \multicolumn{2}{|c|}{\textbf{Extremely Hard Maze}} \\ 
   
    \multicolumn{2}{|c}{($E=25,000$)} & 
    \multicolumn{2}{|c}{($E=25,000$)} & 
    \multicolumn{2}{|c}{($E=75,000$)} & 
    \multicolumn{2}{|c|}{($E=75,000$)} \\
     \hline
    \textbf{Novelty} & \textbf{Surprise} 
    &
    \textbf{Novelty} & \textbf{Surprise} 
    &
    \textbf{Novelty} & \textbf{Surprise} 
    &
    \textbf{Novelty} & \textbf{Surprise} 
    \\
    \hline 
   \parbox[c][0.2in]{0.7in}{\includegraphics[width=0.7in]{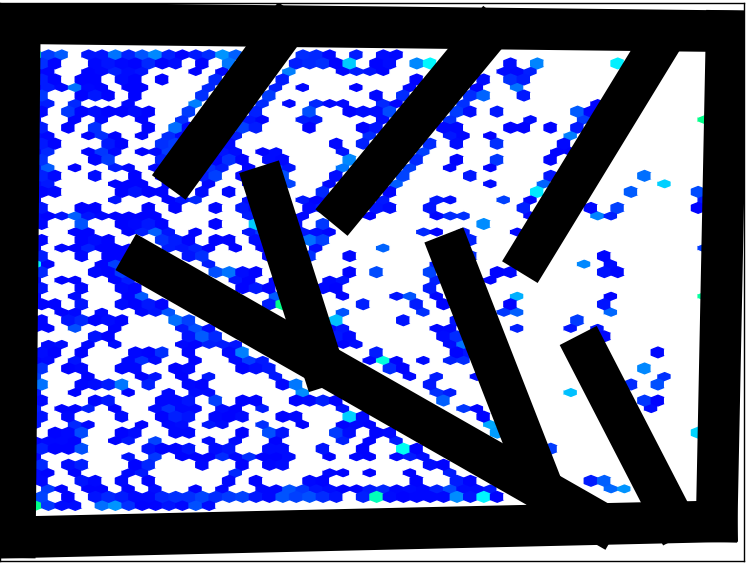}} &
   \parbox[c][0.2in]{0.7in}{\includegraphics[width=0.7in]{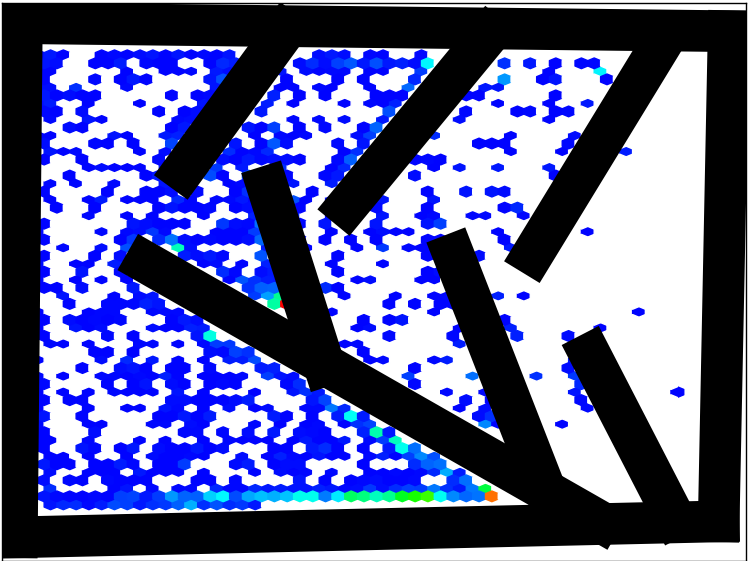}}
    &
   \parbox[c][0.7in]{0.7in}{\includegraphics[width=0.7in]{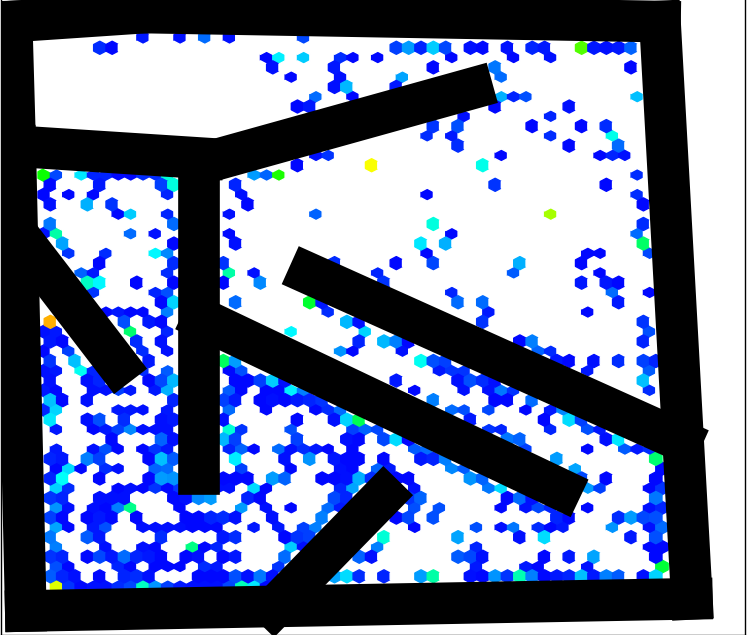}} &
   \parbox[c][0.7in]{0.7in}{\includegraphics[width=0.7in]{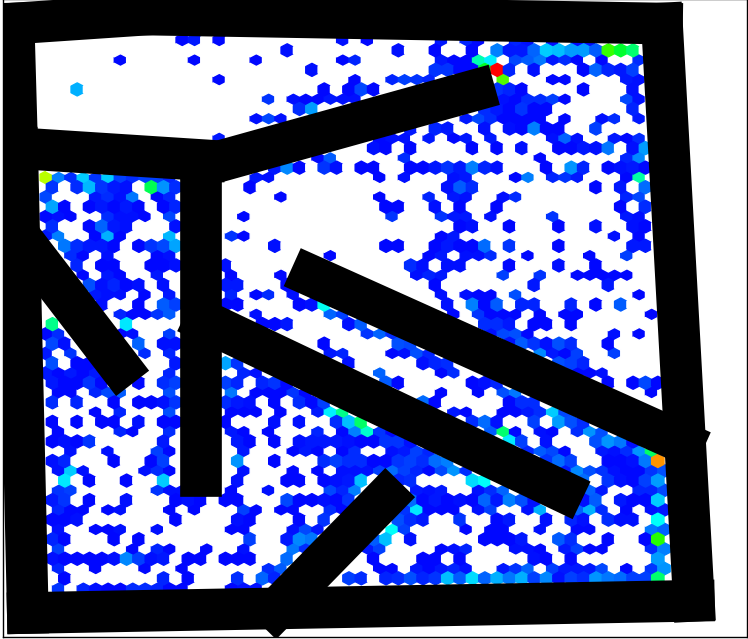}}	
    &
   \parbox[c][0.7in]{0.7in}{\includegraphics[width=0.7in]{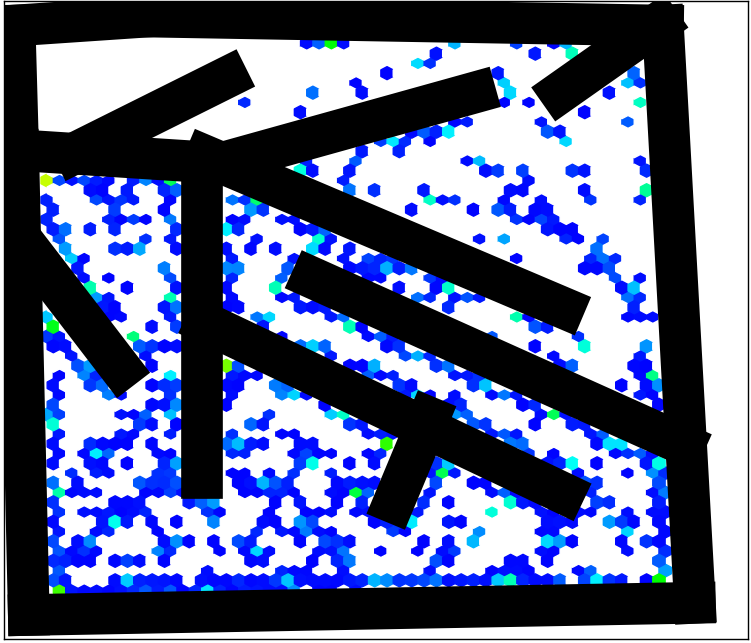}} &
   \parbox[c][0.7in]{0.7in}{\includegraphics[width=0.7in]{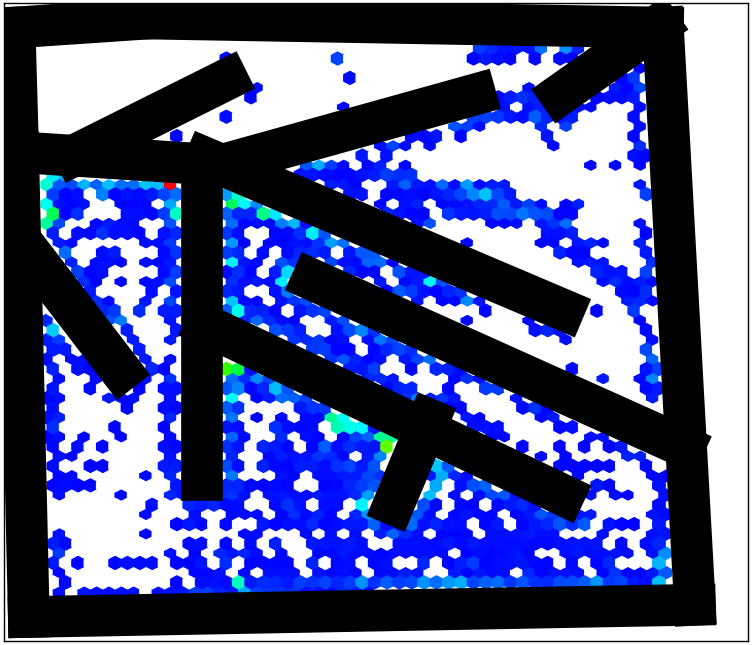}}	
    &
    \parbox[c][0.7in]{0.7in}{\includegraphics[width=0.7in]{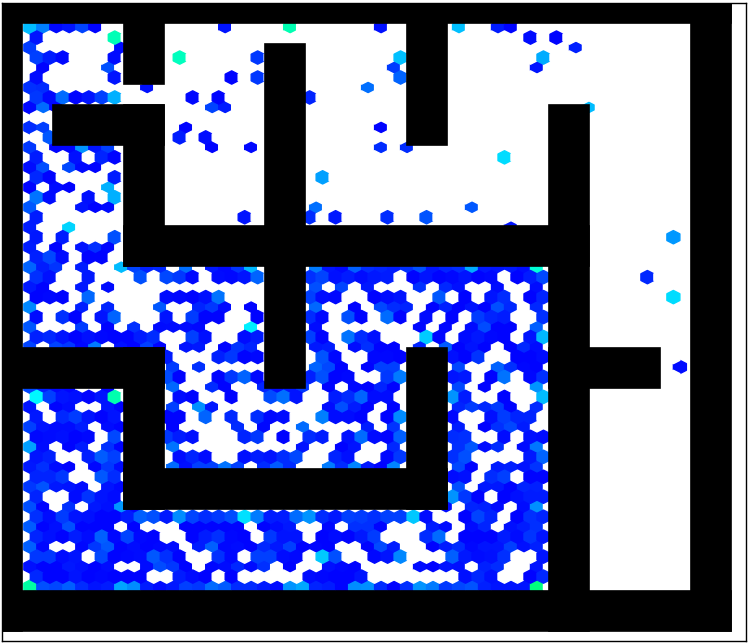}} &
   \parbox[c][0.7in]{0.7in}{\includegraphics[width=0.7in]{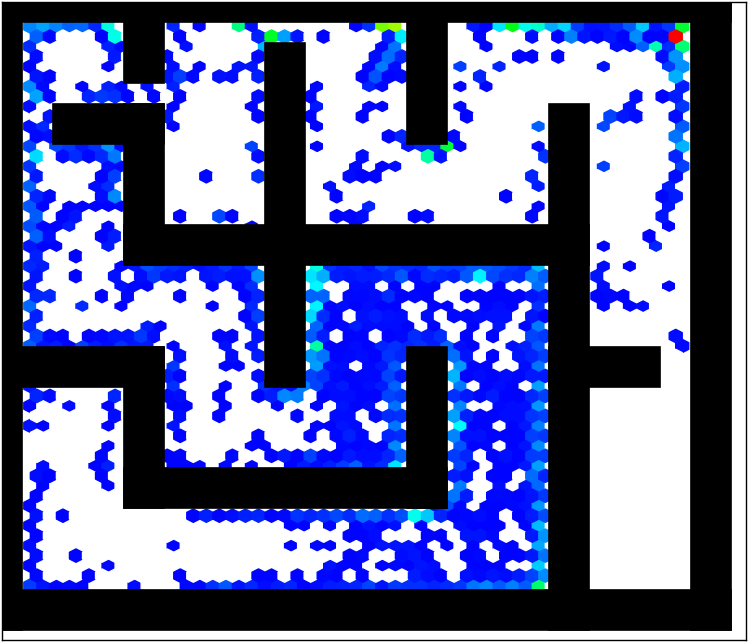}}	
    \\
    
    \multicolumn{2}{|c}{\includegraphics[width=1.5in]{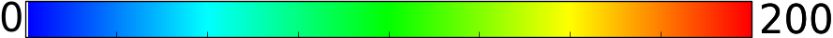}} & 
    \multicolumn{2}{|c}{\includegraphics[width=1.49in]{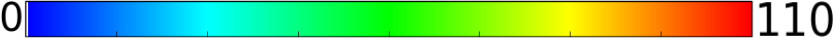}} & 
    \multicolumn{2}{|c}{\includegraphics[width=1.49in]{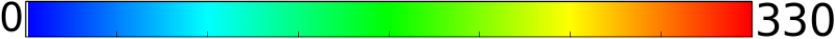}} & 
    \multicolumn{2}{|c|}{\includegraphics[width=1.5in]{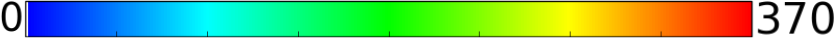}} \\
    
     $H=0.63$ & $H=0.67$ & $H=0.61$ & $H=0.67$ & $H=0.63$ & $H=0.69$ & $H=0.64$ & $H=0.68$\\
    
    \hline
   
  \end{tabular}
  \egroup
\end{table*}

\begin{table*}[!tb]
\centering
\caption{\textbf{Genotypic Space}. Metrics of genomic complexity and diversity of the final ANNs evolved using NEAT, averaged across successful runs. Values in parentheses denote standard deviations.} 
    \label{table:genotypic_metrics}

    \begin{tabular}{|c|c|c|c|c|c|c|c|}
    \hline
    \multirow{2}{*}{\textbf{Maze}} &
    \multirow{2}{*}{\textbf{Algorithm}} &
      \multicolumn{2}{c|}{\textbf{Genomic Complexity}} &
      \multicolumn{4}{c|}{\textbf{Genomic Diversity}} \\
    \cline{3-8}
    & & Connections & Hidden Nodes & Compatibility & Disjoint & Weight Difference & Excess \\
    \hline
    \hline

     \multirow{2}{*}{\textbf{Medium}} & Surprise & 33.76 (15.08) &  2.46 (1.53) & 42.52 (19.80) & 27.40 (16.27) & 1.22 (0.25) & 11.44 (11.45)\\ 
     \multirow{2}{*}{} & Novelty & 29.08 (6.10) & 2.2 (1.0) & 32.55 (7.90) & 24.97 (7.05) & 1.09 (0.26) & 4.28 (3.41) \\
    
    \hline
    \hline

   \multirow{2}{*}{\textbf{Hard}} & Surprise & 52.34 (28.57) & 3.84 (2.66) & 73.24 (36.82) & 51.60 (27.76) & 1.25 (0.24) & 17.86 (20.99)\\
    \multirow{2}{*}{} & Novelty & 32.55 (9.84) & 2.48 (1.29) & 39.35 (12.05) & 31.06 (11.32)& 1.19 (0.28) & 4.71 (4.66) \\

    \hline
    \hline    

    \multirow{2}{*}{\textbf{Very Hard}} & Surprise &  101.94 (52.45) &  6.94 (3.62) & 160.31 (67.01) & 121.56 (57.11) & 1.26 (0.22) & 34.96 (37.34)\\
    \multirow{2}{*}{} & Novelty & 42.03 (14.53) & 3.27 (1.90) & 56.53 (19.66) & 46.65 (19.25) & 1.16 (0.27) & 6.38 (7.71) \\
    \hline
    \hline

    \multirow{2}{*}{\textbf{Extremely Hard}} & Surprise & 158.16 (89.65) &  10.63 (5.85) & 260.52 (113.08) & 207.97 (107.2) & 1.31 (0.23) & 48.62 (54.20)\\
    \multirow{2}{*}{} & Novelty & 41.79 (11.77) & 3.25 (1.53) & 54.05 (14.44) & 45.30 (14.55) & 1.15 (0.24) & 5.28 (4.93) \\
    \hline
  \end{tabular}
\end{table*}
The capacity of surprise search is more evident in the very hard maze (see Fig.~\ref{fig:rob_very_hard}) where the difference in terms of robustness becomes even larger between surprise and novelty search. While in the first \num[group-separator={,}]{50000} evaluations novelty and surprise search attain a comparable number of successes, the performance of surprise search is boosted for the remainder of the evolutionary run. Ultimately, surprise search solves the very hard maze in 99 out of 100 times in just \num[group-separator={,}]{160000} evaluations whereas novelty search manages to obtain 85 solutions by the end of the \num[group-separator={,}]{250000} evaluations. With 88 solutions out of 100, SS$_{np}$ performs similarly to novelty search in this maze but is generally slower compared to surprise search. Objective search and the two random baselines, as expected, do not succeed in solving the maze.

Similarly, in the extremely hard maze (see Fig.~\ref{fig:rob_extreme_hard}) the benefits of surprise search over the other algorithms are quite apparent. While surprise and novelty obtain a similar number of successes in the first \num[group-separator={,}]{100000} evaluations, surprise search obtains more successes in the remaining evaluations of the run. At the end of \num[group-separator={,}]{250000} evaluations in the most deceptive map examined, surprise search finds solutions in 67 runs versus 48 runs of novelty search. SS$_{np}$ finds 39 solutions and it is generally slower than novelty and surprise search. As in the very hard maze, the remaining algorithms fail to find a single solution to this maze.

\subsection{Analysis}
\label{subsec:analysis}

\newadditions{
As an additional comparison between surprise and novelty search, we study the behavioural and genotypical characteristics of these two approaches. The behavioural space is presented in a number of typical runs collected from the four mazes, while the genotypical space is inspected through the metrics computed from the final ANNs evolved by these two algorithms.
} 
Objective search and the other baselines are not further analysed in this section to emphasise on comparisons between surprise and novelty search.

\subsubsection{Behavioural Space: Typical Examples}

Table~\ref{fig:heatmaps} shows pairs of typical surprise and novelty search runs for each of the four mazes; in all examples illustrated the maze is solved at different number of evaluations as indicated at the captions of the images.
The typical runs are shown as heatmaps which represent the aggregated distribution of the robots\rq{} final positions throughout all evaluations. Moreover, we report the entropy ($H$) of those positions as a measure of the populations\rq{} spatial diversity in the maze.
\newadditions{Surprise search seems to explore more uniformly the space, as revealed by the final positions depicted in the heatmaps. The corresponding $H$ values further support this claim, especially in the more deceptive mazes.
}

\subsubsection{Genotypic Space}

Table \ref{table:genotypic_metrics} contains a set of metrics that characterize the final ANNs evolved by surprise and novelty search obtained from all four mazes, which quantify aspects of genomic \emph{complexity} and genomic \emph{diversity}. For genomic complexity we consider the number of connections and the number of hidden nodes of the final ANNs evolved, while genomic diversity is measured as the average pairwise distance of the final ANNs evolved. This distance is computed with the compatibility metric, a linear combination of disjoint and excess genes and weight difference, as defined in \cite{stanley2002neat}. As noted in \cite{lehman2011abandoning}, novelty search tends to evolve simpler networks in terms of connections when compared to objective search. Surprise search, on the other hand, seems to generate significantly more densely connected ANNs than novelty search (based on the number of connections). It also evolves slightly larger ANNs than novelty search (based on the number of hidden nodes). Most importantly, surprise search yields population diversity --- as expressed by the compatibility metric \cite{lehman2011abandoning} --- that is significantly higher than novelty search. This difference seems to be mostly due to the disjoint factor, which counts the number of mismatching genes between two genomes, depending on whether their genes are within the innovation numbers of the other genome \cite{stanley2002neat}. This suggests that ANNs evolved with surprise search are more diverse in terms of evolutionary history. 
In the more deceptive mazes, differences in genomic complexity and diversity become significantly larger. In the very hard maze the average number of connections for surprise search grows to $101.94$ ($\sigma = 52.45$) while novelty search evolves ANNs with $42.03$ connections ($\sigma =14.53$), on average; the number of hidden nodes used by surprise search is significantly larger ($6.94$; $\sigma = 3.62$) compared to novelty search. Moreover the diversity metric (compatibility) is around three times that of novelty search. A similar trend can be noticed in the extremely hard maze, where again surprise search evolves denser, larger and more diverse ANNs. As mentioned earlier, handling more complex and larger ANNs has a direct impact on the computational cost of surprise search since it takes more time to simulate new networks across generations. It should be noted that creating larger networks does not imply that this behaviour is beneficial, it is however an indication that surprise search operates differently to novelty search.

\section{Surprise Search in Generated Mazes}\label{experiments_maze_generation}

In the previous sections we showed the power of surprise search in four selected instances of deceptive problems. While surprise search outperforms novelty and objective search both in terms of efficiency and robustness in four human-designed mazes, an important concern is whether these results are general enough across a broader set of problems.

In order to assess how surprise search generalises in any maze navigation task, we follow the methodology presented in \cite{lehman2011novelty} and test the performance of surprise, novelty and objective search as well as the baselines across numerous mazes generated through an automated process.
Moreover, the parameters of $k$ and $n$ which were fine-tuned for the problem at hand in each maze of Section \ref{sec:experiments} are now kept the same, enabling us to observe if a particular parameter setup for surprise search can perform well in unseen problems of varying complexity. 

\subsection{Experiment Description}

\begin{figure}[t]
\centering
\captionsetup[subfigure]{labelformat=empty}
\subfloat[2 subdivisions]{
\includegraphics[width=0.17\linewidth]{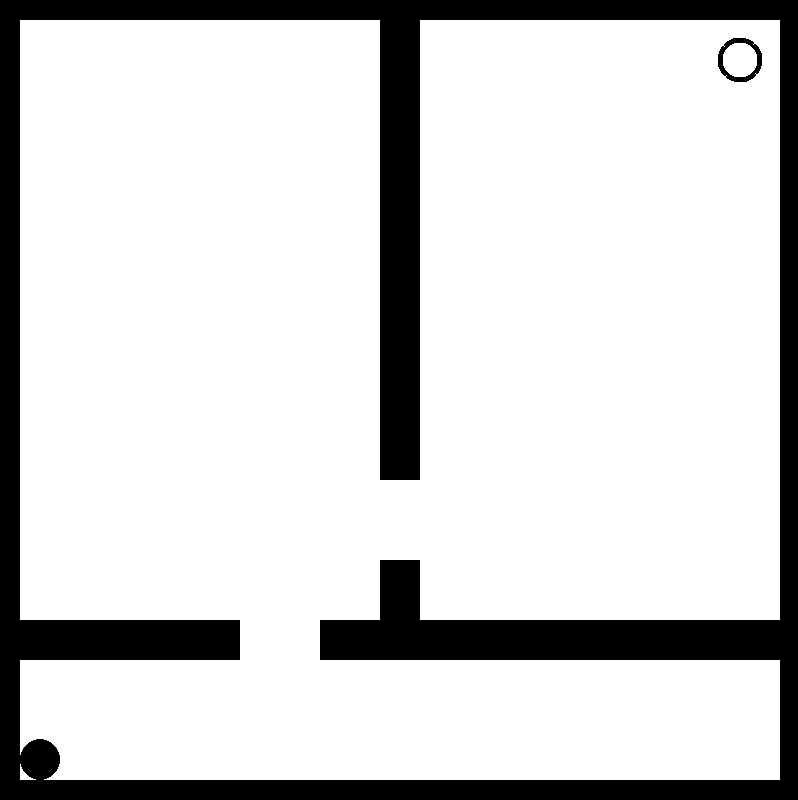} \label{fig:maze_first}}
\subfloat[3 subdivisions]{
\includegraphics[width=0.17\linewidth]{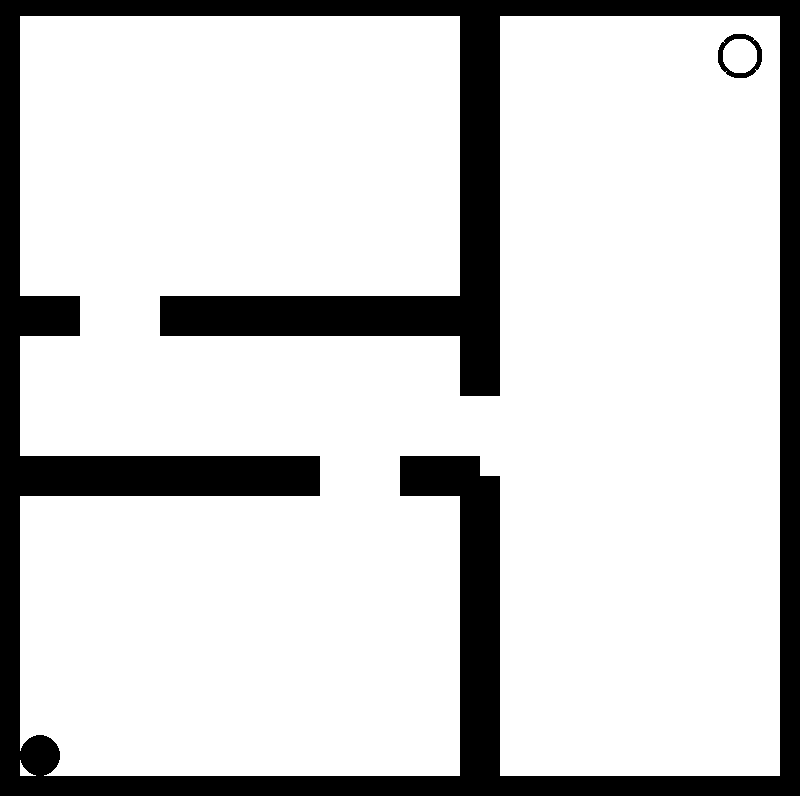} \label{fig:maze_second}}
\subfloat[4 subdivisions]{
\includegraphics[width=0.17\linewidth]{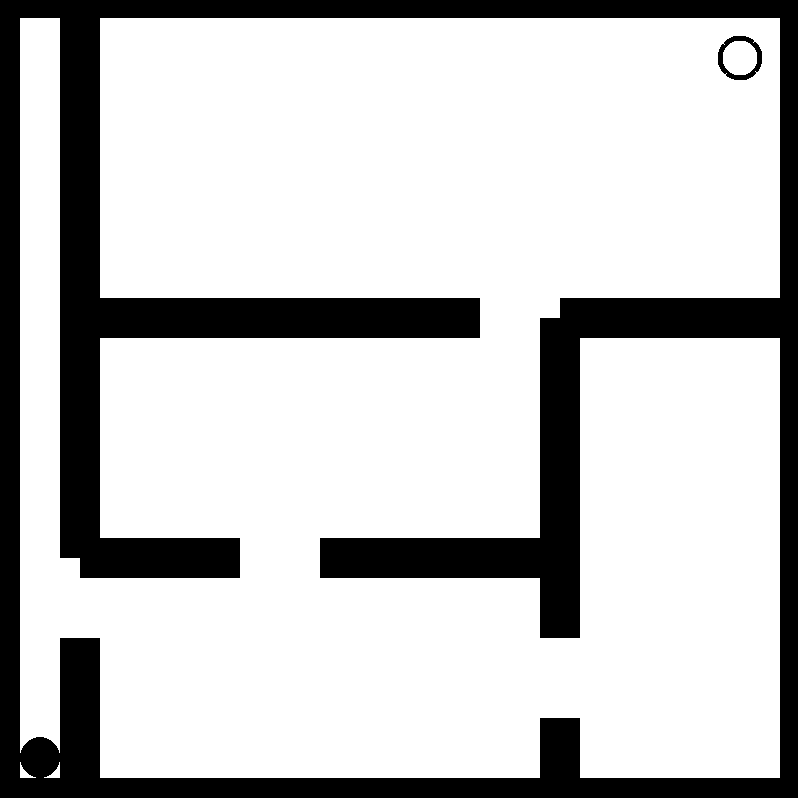} \label{fig:maze_third}}
\subfloat[5 subdivisions]{
\includegraphics[width=0.17\linewidth]{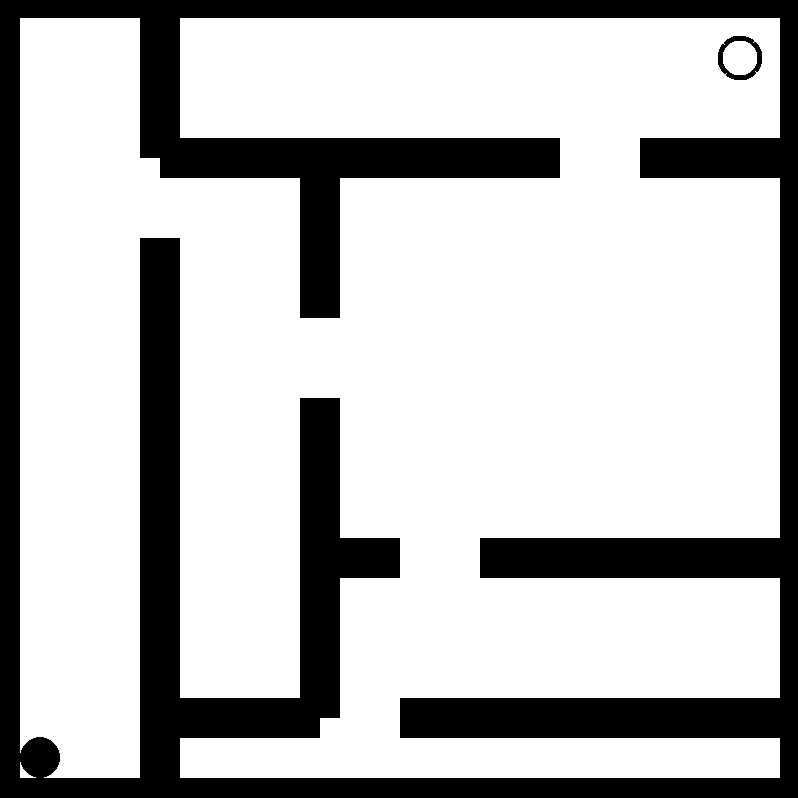} \label{fig:maze_fourth}}
\subfloat[6 subdivisions]{
\includegraphics[width=0.17\linewidth]{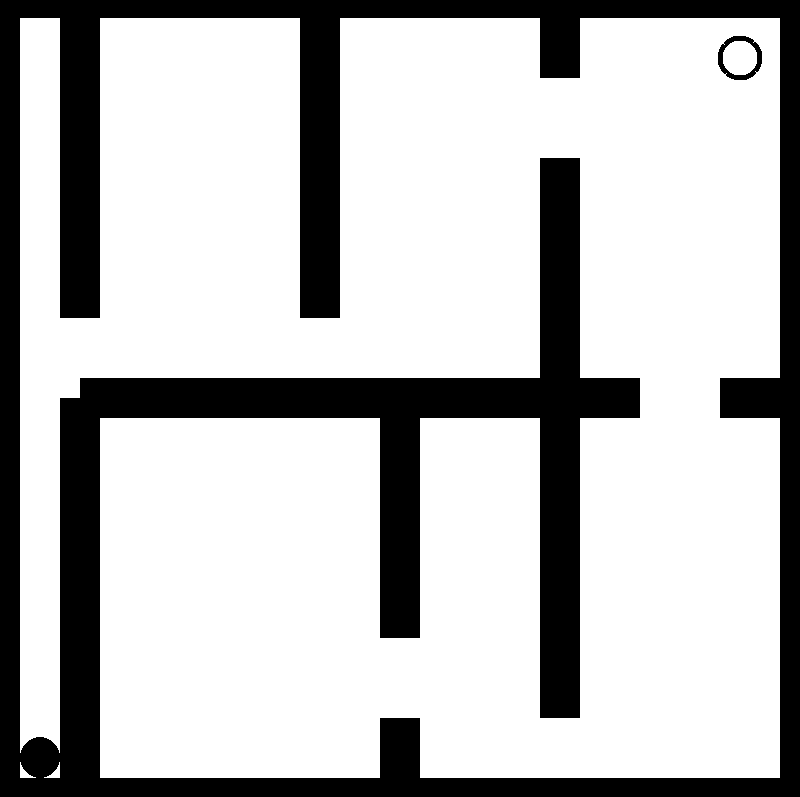} \label{fig:maze_five}}
\caption{\textbf{Maze generator:} Sample generated mazes (200x200 units) created via recursive division, showing the starting location (black circle) and the goal location (white circle).}
\label{fig:maze_examples}
\end{figure}

To compare the capabilities in navigation policies of surprise, novelty and objective search in increasingly complex maze problems, we test their performance against 60 randomly generated mazes. These mazes are created by a recursive division algorithm \cite{reynolds2010maze}, which starts from an empty maze and divides it into two areas by adding a vertical or a horizontal wall with a randomly located hole in it. This process is repeated until no areas can be further subdivided, because doing so would make the maze untraversable or because a maximum number of subdivisions is reached. In this experiment, the starting position and the ending position of the maze have been fixed in the lower left and upper right corner respectively, while the generated mazes have a number of subdivisions chosen randomly between 2 and 6. These values have been chosen empirically to avoid generating mazes that are too easy (solvable by all three methods in few generations) or impossible to solve (because of too many subdivisions). Examples of the mazes generated are shown in Fig. \ref{fig:maze_examples}.
The parameters of surprise search and novelty search are fixed based on well-performing setups with mazes of Section \ref{sec:experiments}: surprise search uses $k = 200$ and $n = 2$ (used in the very hard maze) and novelty uses $n=15$ (used in medium, hard and very hard mazes). Each generated maze was tested 50 times for each of three methods, measuring the number of successes (i.e.~once the agent reaches the goal) in each maze. The number of simulation timesteps is set to $200$ and the number of generations to $600$.

\subsection{Results}

As a first analysis on the results obtained on the 60 mazes, we focus on which of the evolutionary approaches finds strictly more successes. Table \ref{table:rankings} shows that surprise search has more successes than novelty search in 40\% of mazes, while novelty achieves more successes than surprise search in 8\% of the generated mazes. Comparing the results of these two approaches against objective search, surprise search outperforms objective search in more mazes (56\%) than novelty search (40\%).
\newadditions{
If we look at the baselines, $SS_{np}$ reaches comparable performance to novelty search, but surprise search remains the most successful algorithm, as it outperforms $SS_{np}$ in 45\% of the considered mazes.
Finally, $SS_{r}$ and random search evidently perform poorly compared to the other approaches.
}

\begin{figure}[!tb]
\centering
\begin{minipage}{0.22\textwidth}
\centering
\includegraphics[height=0.6\linewidth]{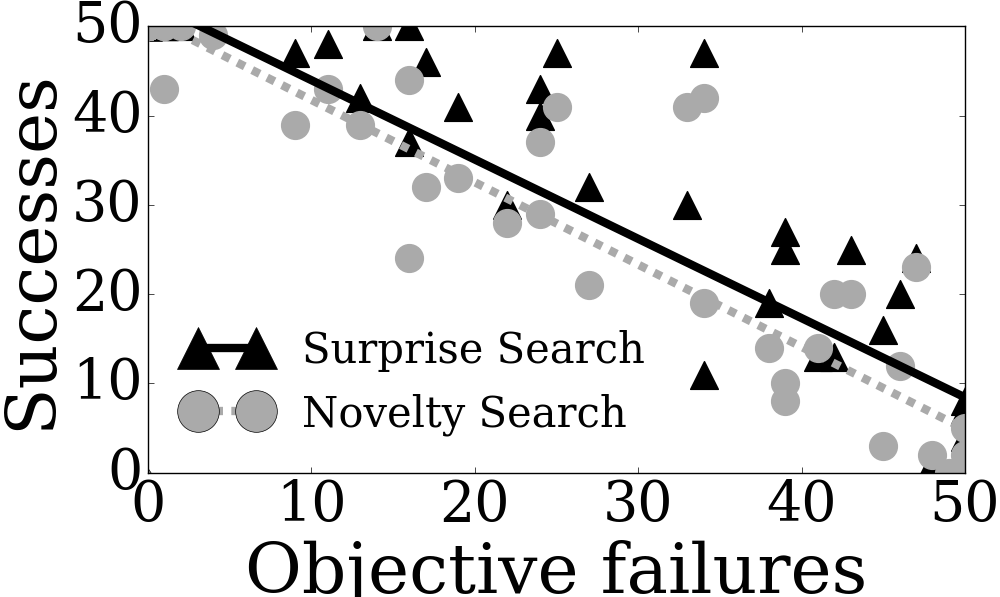}
\caption{\textbf{Linear regression:} relation between the failures of objective search  and successes of surprise and novelty search.}
\label{fig:regression}
\end{minipage}
~
\begin{minipage}{0.22\textwidth}
\centering
\includegraphics[height=0.6\linewidth]{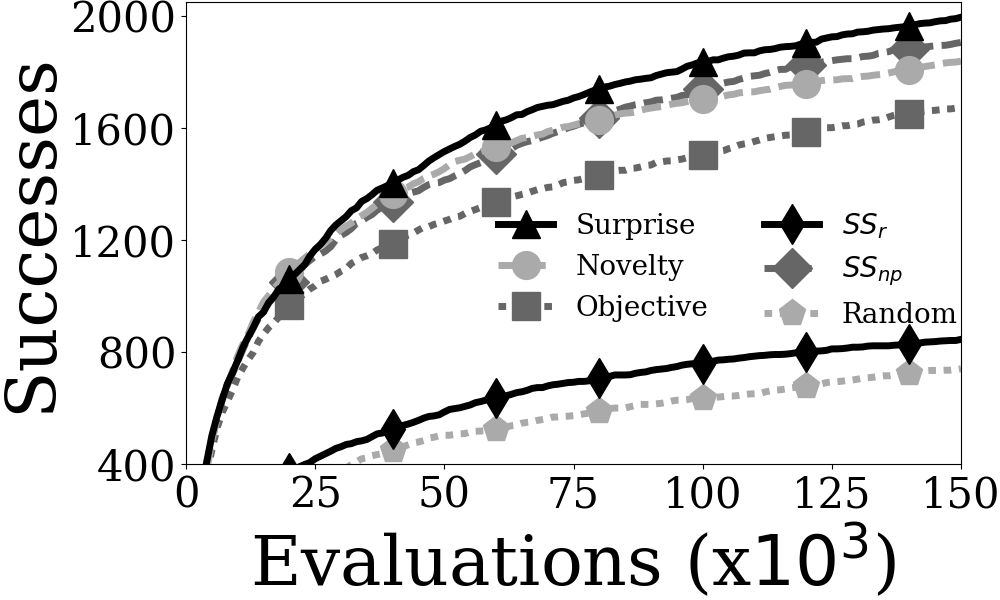}
\caption{\textbf{Robustness:} algorithm successes in solving all the generated mazes over the number of evaluations for each considered method.}
\label{fig:robustness_all_run}
\end{minipage}
\end{figure}

An important question to put to the test is how novelty search and surprise search perform with respect to maze deceptiveness. 
Intuitively, we can say that the deceptiveness (or difficulty) of the maze can be determined by the number of successes obtained by objective search: the more deceptive the maze, the more often objective search would fail to find a solution.
Figure \ref{fig:regression} shows the number of successes obtained by novelty and surprise search against the number of failures obtained by objective search. 
Unsurprisingly, surprise search and novelty search find mazes where objective search failed more difficult as well, since their successes are highly correlated with the failures of objective search (adjusted $R^{2} > 0.85$ for each method, $p < 0.001$).
Looking at the trends of the linear regression lines, surprise search constantly achieves a greater number of successes, based on the intercept values of the linear regression models which are significantly different according to an ANCOVA test ($p < 0.05$). Based on the angle of the linear regression line, it also seems that surprise search scales better for more deceptive mazes.

\begin{table*}[!tb]
\centering
\caption{\textbf{Successes:} Percentage of generated mazes for which the algorithm in the row has a strictly greater number of successes than the algorithm in the column. The last row and the last column are respectively the average of each column and the average of each row.}\label{table:rankings}
\begin{tabular}{|l||c|c|c|c|c|c|c|}
\hline    
                                   & Objective & Novelty & Surprise  & $SS_{np}$ & $SS_{r}$ & Random & Total\\
\hline
\hline
Objective                      & - & 15 & 5 & 6 & 71 & 78 & 35 \\     
\hline
Novelty                         & 40 & - & 8 & 20 & 75 & 81 & 44.8 \\
\hline    
Surprise          &  56 &  40 & - & 45 & 78 & 85 & \textbf{60.8} \\   
\hline
$SS_{np}$                  & 51 & 35 & 11 & - & 78 & 85 & 52 \\     
\hline
$SS_{r}$                      & 1 & 1 & 1 & 1 & - & 36 & 8\\     
\hline
Random                        & 0 & 0  & 0 & 0 & 20 & - & 4\\     
\hline
Total                            & 29.6 & 19  & \textbf{5} & 16.5 & 62.8 & 71.8 & -\\     
\hline
\end{tabular}
\end{table*}
As a final analysis, we report the robustness obtained by aggregating all the runs of the 60 generated mazes for each approach, i.e. a total of 3000 runs. From Fig. \ref{fig:robustness_all_run} we can observe that  surprise search is faster, on average, than novelty and objective search in reaching the goal from $112,500$ evaluations onward ($p < 0.05$). Surprise, novelty and objective search require on average $71,865$ ($\sigma  = 63,007$), $76,225$ ($\sigma = 64,961$) and $84,398$ ($\sigma = 65,081$) evaluations for each success, respectively. 
As Fig.~\ref{fig:robustness_all_run} shows, some maze problems are easy to solve as all three methods fare similarly in the first $20,000$ evaluations but as the problems become more complex, surprise and novelty search become faster than objective search and eventually surprise search surpasses novelty search in terms of successes.
\newadditions{
Furthermore, surprise search shows a significant improvement compared to its baseline variants. Respectively, $SS_{np}$, $SS_{r}$ and random search obtain $75,531$ ($\sigma  = 63,820$), $118,668$ ($\sigma  = 53934$) and  $123,558$ ($\sigma  = 50,568$) evaluations; all results are significantly different from the performance of surprise search ($p < 0.05$).
}

\section{Discussion}\label{sec:discussion}

This paper identified the notion of \emph{surprise}, i.e. deviation from expectations, as an alternative measure of divergence to the notion of novelty and presented a general framework for incorporating surprise in evolutionary search in Section \ref{sec:surprisesearch}. In order to highlight the differences between surprise search and other divergent search techniques (such as novelty search) or baselines (such as random search or search with inaccurate predictions), an experiment in the robot maze navigation testbed was carried out and comparisons between algorithms were made on several dimensions. The key findings of these experiments suggest that surprise search yields comparable efficiency to novelty search and it outperforms objective search. Moreover it finds solutions faster and more often than any other algorithm considered. In a broader range of mazes, generated via recursive subdivision, surprise search was also shown to be more robust and generalise well, as it had more successes than novelty search in 40\% of generated mazes.

The comparative advantages of surprise search over novelty search are inherent to the way the algorithm searches, attempting to deviate from predicted \emph{unseen} behaviours instead of prior \emph{seen} behaviours. Compared to novelty search, surprise search may also deviate from expected behaviours that exist in areas that have been visited in the past by the algorithm. The novelty archive operates in a similar fashion; however it contains positions (instead of prediction points) and these positions are always considered for the calculation of the novelty score. In surprise search, instead, the prediction points are derived from clusters that characterise \emph{areas} in the behavioural space. 

\newadditions{
The findings in the maze navigation experiments show a clear difference between novelty and surprise, both behaviourally and genotypically. Surprise search has greater exploratory capabilities, which are more obvious in later generations. Surprise search also creates genotypically diverse populations, with larger and denser ANNs. It is likely this combination of diverse populations, larger and denser networks and a higher spatial diversity that gives surprise search its advantage over novelty search.
}

Finally, the comparative analysis of surprise search against random search suggests that surprise search is not random search. Clearly it outperforms random search in efficiency and robustness. Furthermore, the poor performance of the two surprise search variants --- employing random predictions and omitting predictions --- suggests that the prediction of expected behaviour is beneficial for divergent search.

By now we have enough evidence for the benefits of surprise search and enough findings suggesting that surprise search is a different and more robust algorithm compared to novelty search in this domain. Furthermore, through our analysis, we have identified qualitative characteristics of the algorithm that gave us critical insights on the way the algorithm operates. However, we still lack empirical evidence on the reasons the algorithm manages to perform that well compared to other divergent search algorithms. An intuition by the anonymous reviewers of a previous paper about the comparative benefits of surprise search is that the algorithm allows search to \emph{revisit} areas in the behavioural space. Such a behaviour, in contrast, is penalised in novelty search. This difference in how the two algorithms operate leads to the assumption that surprise search is more willing to revisit points in the behaviour space --- in a form of \emph{backtracking} or cyclical manner. As a result of this shifting selection pressure in the behavioural space a different strategy is adopted every time a particular area is revisited as each time the ANN controller is different and potentially larger. Such an algorithmic behaviour appears to be beneficial for search and might explain why ANNs get significantly larger in surprise search. 

More importantly than a performance comparison between algorithms, however, is the introduction of surprise as a drive for divergent search. As will be discussed in Section \ref{sec:futurework}, the general framework of surprise search can be used with other behaviour characterizations and in other domains. Moreover, while some properties such as the model of prediction are inherent to surprise search, properties such as the clustering of behaviours as well as findings regarding the ability of surprise search to backtrack can be re-used in other divergent search algorithms; examples include a variant of novelty search which considers neighboring cluster centroids rather than neighboring individuals in the behavioural space, or a novelty archive which is pruned (and reduces in size) over time to allow backtracking. We can only hypothesise that the way surprise search operates may result in increased \emph{evolvability} \cite{lehman2011improving}, which is an individual's capacity to generate future phenotypic variation, or alternatively, the potential for further evolution. Naturally, all these hypotheses need to be tested empirically in future studies as outlined in the next section.

\section{Extensions and Future Work}
\label{sec:futurework}

While this study already offers evidence for the advantages of surprise as a form of divergent search, further work on several directions needs to be performed. 

\begin{figure}[!tb]
\captionsetup[subfigure]{labelformat=empty}
\centering
\subfloat[Generation $t - 2$]{
\includegraphics[width=0.3\linewidth]{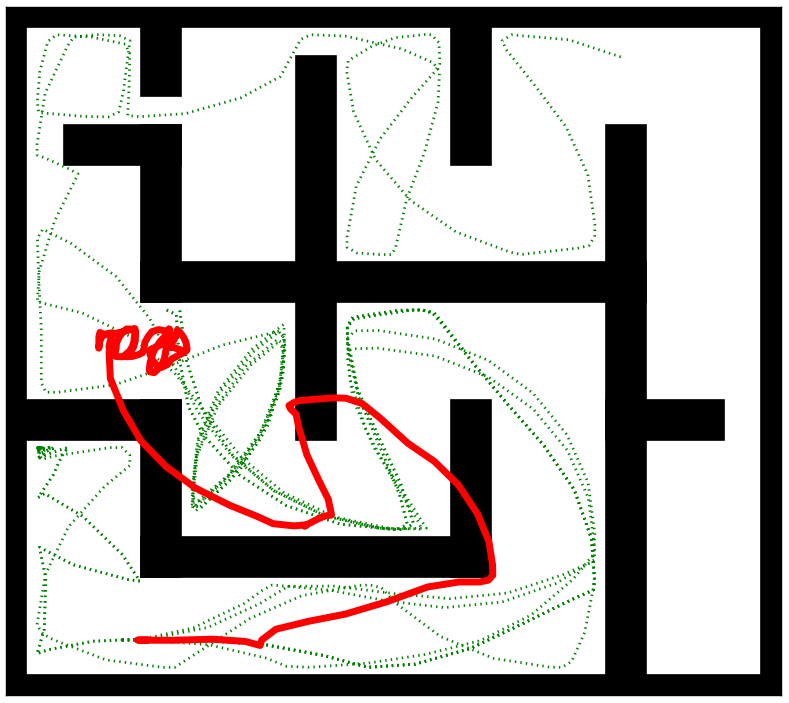} \label{fig:BC_first}}
\subfloat[Generation $t - 1$]{
\includegraphics[width=0.3\linewidth]{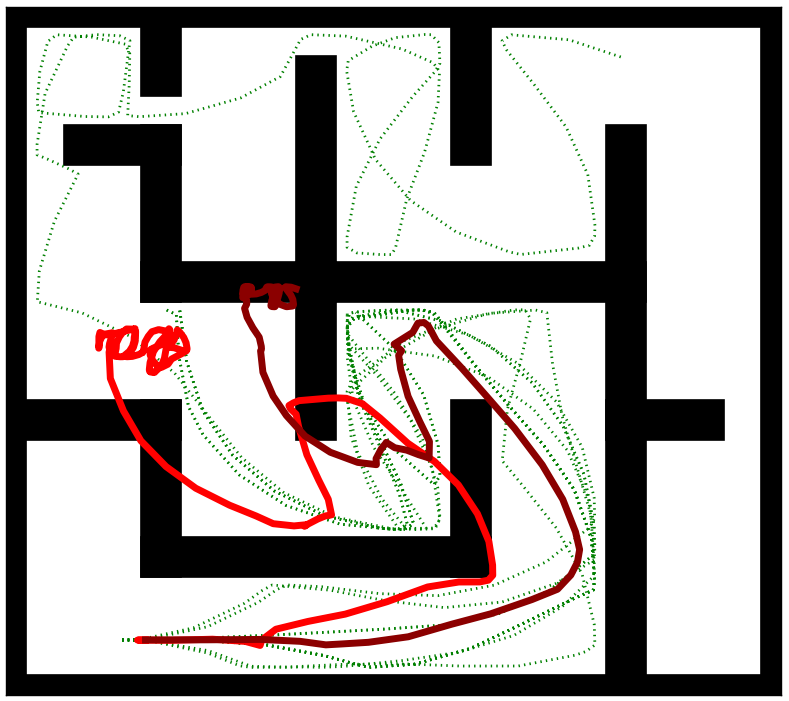} \label{fig:BC_second}}
\subfloat[Generation $t$]{
\includegraphics[width=0.3\linewidth]{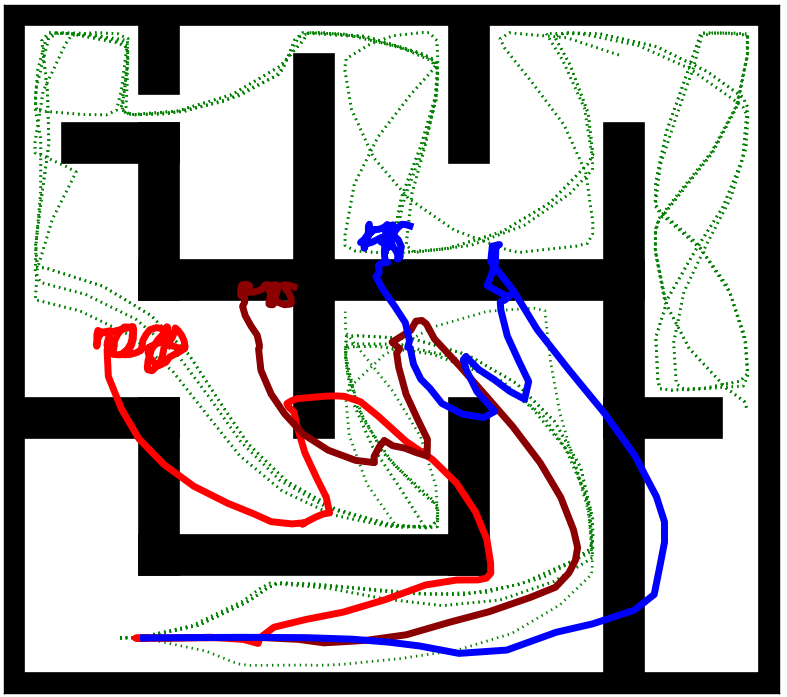} \label{fig:BC_third}}
\caption{\textbf{Behaviour Characterization:} The key phases of the surprise search algorithm on the maze navigation task, characterizing behaviour via 200 samples of robot positions over time. Surprise search uses a history of two generations ($h=2$) and 10 clusters ($k=10$) in this example (for the sake of visualization, only one cluster is shown). Robot trails are depicted as green lines. Cluster centroids in generations $t-2$ and $t-1$ as well as their predictions are depicted, respectively, as red, dark red and blue lines.}
\label{fig:BC_surprise_search}
\end{figure}

\textbf{Behaviour Characterization:} The surprise search algorithm currently characterises a behaviour merely as a point (i.e. the final robot position on the maze after the simulation time elapses). While such a decision was made in order to compare our findings against the initial results of \cite{lehman2011abandoning}, we currently have no evidence suggesting that surprise search would be able to generalise well in behaviours that are characterised by higher dimensions (e.g. a robot trail). To envision the behaviour of surprise search in a high dimensional space, Fig. \ref{fig:BC_surprise_search} shows a possible implementation for surprise search with robot trails, sampled over time. Following the implementation described in Section \ref{sec:experiments_surprise}, we show the three key phases of the algorithm: the robots' trails are clustered at generation $t-2$ via $k$-means (see Fig. \ref{fig:BC_first}), then at generation $t-1$ we seed the clustering algorithm with the ones computed in the previous generation and we find the new trajectories' centroids (Fig. \ref{fig:BC_second}). Finally the prediction at generation $t$ is computed via linear interpolation of each point on the computed (centroid) trajectories at generations $t-2$ and $t-1$ (Fig. \ref{fig:BC_third}). Preliminary experiments have shown that predictions of robot trails do not affect the performance of surprise search compared to results in this paper; future studies, however, should investigate the impact of higher dimensional behaviours across several domains, especially on how they affect the predictive model of the algorithm.

\begin{figure}[!tb]
\captionsetup[subfigure]{labelformat=empty}
\centering
\subfloat[$H_{t-2}$ at generation $t - 2$]{
\includegraphics[width=0.3\linewidth]{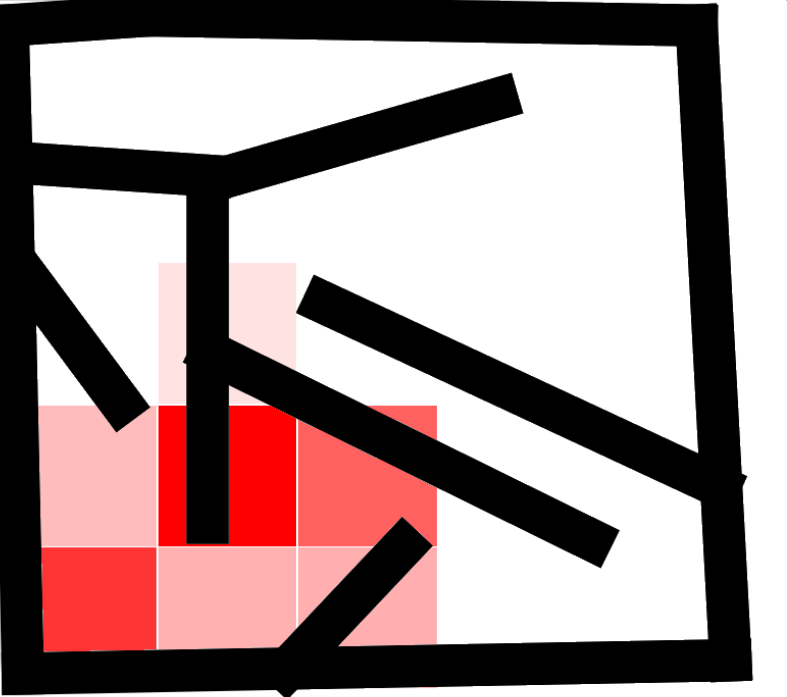}}~
\subfloat[$H_{t-1}$ at generation $t - 1$]{
\includegraphics[width=0.3\linewidth]{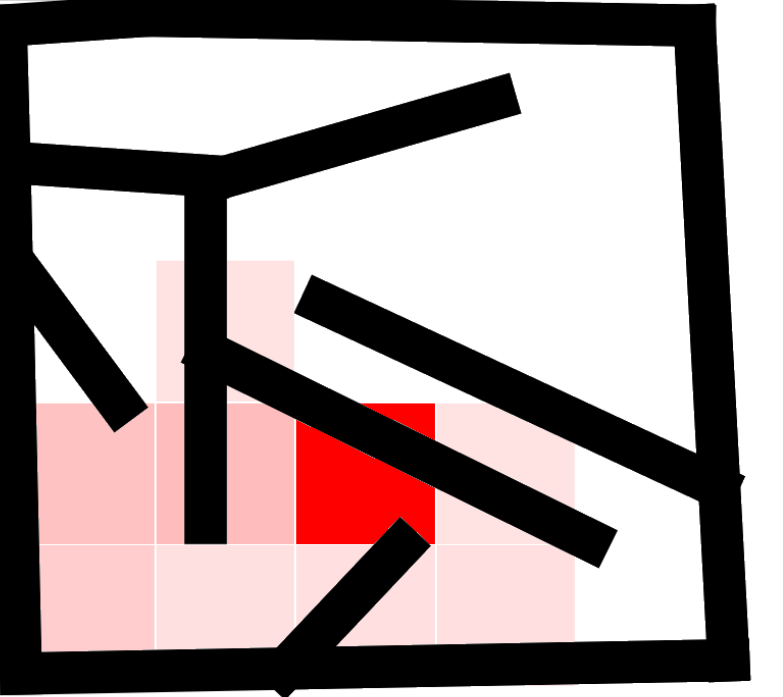}}~
\subfloat[$H_{t}$ at generation $t$]{
\includegraphics[width=0.3\linewidth]{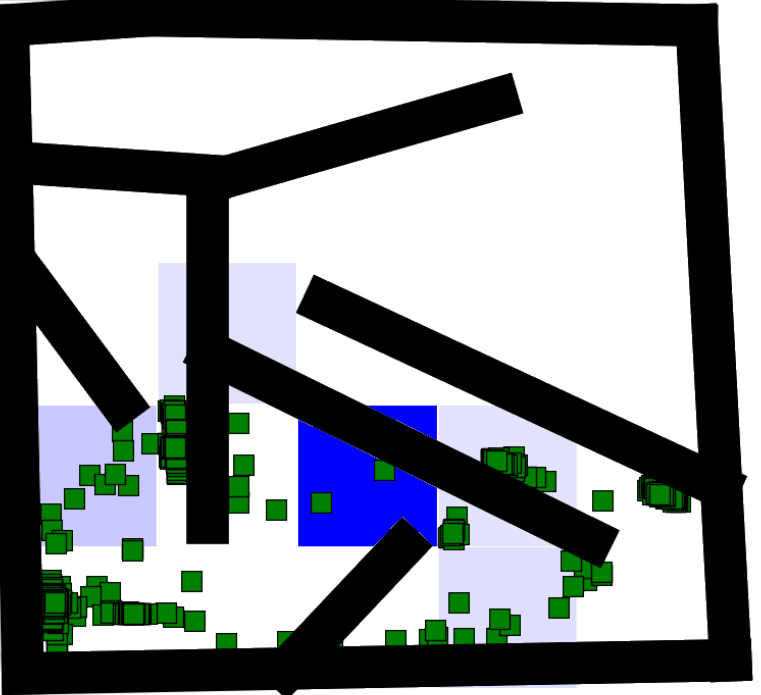}}
\caption{\textbf{Deviation:} Surprise search using heatmaps, at generation $t$. The first two heatmaps are computed in the last two generations by using the final robot positions, $H_{t-2}$ and $H_{t-1}$. Using linear interpolation, the difference $H_{t-1} - H_{t-2}$ is computed and applied to $H_{t-1}$ to derive the predicted current population's $H_t$. The surprise score penalizes a robot if its position (green point) is on a high concentration cell on the predicted heatmap $H_t$.}
\label{fig:surprise_heatmaps}
\end{figure} 

\textbf{Deviation:} The current algorithm allows for any degree and type of deviation from the expected behaviour. Inspired by novelty search, this paper only investigated a linear deviation from expectations --- i.e. the further a behaviour is from the prediction the better. There exist, however, several ways of computing deviation in a non-linear or probabilistic fashion, e.g. as per \cite{grace2015modeling}. 
In a maze navigation environment for instance, we can alternatively consider a non-distance-based deviation by using heatmaps of the chosen behaviour characterization. Figure \ref{fig:surprise_heatmaps} shows the key phases of this implementation: in generation $t-2$ and $t-1$ we compute the heatmaps $H_{t-2}$ and $H_{t-1}$ which map the final positions of all robots ($k=1$), and we use a linear interpolation to compute the predicted heatmap at generation $t$. The surprise score is then computed by mapping the individual's position on the predicted heatmap: $1 - H_{t}(x,y)$, where  $x, y$ is the final position of the robot mapped onto the heatmap.

\textbf{Complexity and generality:} To a degree, the experiment with 60 generated mazes tests how generalizable surprise search is without explicit parameter tuning. Since results from that experiment indicated that surprise search scales better to more deceptive problems, we need to further test the algorithm\rq{}s potential within the maze navigation domain through more deceptive and complex environments. The capacity of surprise search will then need to be tested in other domains such as robot locomotion or procedural content generation. As an example, the potential of surprise search has been explored for the generation of unexpected weapons in a first person shooter game \cite{gravina2016constrained}. In that work, the considered behaviour characterization is a weapon tester agent's death location: as we have described above, we can employ a heatmap as a probability distribution of the population's behaviour and predict the next generation's heatmap by means of linear interpolation of each singular cell. Surprise search has shown its capacity to generate feasible and diverse content, thereby achieving quality diversity.

\newadditions{
Surprise search has also been successfully implemented to evolve soft robot morphologies \cite{gravina2017softrobots} constrained by a fixed lattice. The goal of this experiment is to create robots able to travel as far as possible from a starting position, within a number of simulation steps. In this task, surprise search was shown to be as efficient as novelty search; moreover, it evolved more diverse morphologies. When computing behavioural distance for this problem, the surprise score is computed by predicting an entire robot trace in all simulation steps, based on past behaviours (traces). This further supports the claim that surprise is unaffected by the dimensionality of the behaviour characterization. The experiments presented in this paper, in \cite{gravina2016constrained} and \cite{gravina2017softrobots} already demonstrate the generalizability of surprise search across three rather diverse domains: maze navigation, game content generation and robot design.
}

\textbf{Model of expected behaviour:} When it comes to designing a model of expected behaviour there are two key aspects that need to be considered: \emph{how much prior information the model requires} and \emph{how is that information used to make a prediction}. In this paper the prediction of behaviour is based on the simplest form of 1-step predictions via a linear regression. This simple predictive model shows the capacity of surprise search given its performance advantages over novelty search. However we can envision that better results can be achieved if machine learned or non-linear predicted models are built on more prior information ($h > 2$). A possible way of considering more extensive history is to apply linear interpolation of past centroids over time. It is thus possible to compute a 3-dimensional line over the three dimensions considered ($x, y, t$) and compute the next predicted position over the line by taking the interpolated position at generation $t$. Linear regression can be easily replaced by a quadratic or cubic regression or even an artificial neural network model or a support vector machine.

\textbf{Prediction locality of surprise search:} The algorithm presented in this paper allows for various degrees of \emph{prediction locality}. We define prediction locality as the amount of local information considered by surprise search to make a prediction. This is expressed by $k$ which is left as a variable for the algorithm designer. Prediction locality can be derived from the behavioural space (as in this paper) but also on the genotypic space. Future investigations should investigate the effect of locality for surprise search. Experiments in this paper (see Fig.~\ref{fig:surprise_sensitivity}) already showcase that algorithm performance is not sensitive with respect to $k$ (as long as $k$ is sufficiently high for the problem at hand).  

\textbf{Clustering:} Surprise search, in the form presented here, requires some form of behavioural clustering. While $k$-means was investigated in the experiments of this paper for its simplicity and popularity, any clustering algorithm is applicable. Comparative studies between approaches need to be investigated, including different ways of dealing with (or taking advantage of) empty clusters.

\section{Conclusions}

\newadditions{
In this paper, we argue that surprise is a concept that can be exploited for evolutionary divergent search, we provide a general definition of the algorithm that follows the principles of searching for surprise and we test the idea in a maze navigation task.
Results show that surprise search has clear advantages over other forms of evolutionary divergent search, i. e. novelty search, and outperforms traditional fitness search in deceptive problems. In particular, surprise search has shown a comparable efficiency to novelty search and, most importantly, to be more successful and faster in finding the goal. Moreover, a detailed analysis of the behaviours and the genomes evolved by surprise search has revealed a more diverse population and a higher exploratory capacity.
}
Finally, the capacity of surprise search to generalize in tasks of increasing complexity is evidently higher when surprise drives the search process, as tested in randomly generated mazes of increasing deceptive properties. These findings support the idea that deviation from expected behaviours can be a powerful alternative to divergent search with key benefits over novelty or objective search. 

\section*{Acknowledgment}

This work has been supported in part by the FP7 Marie Curie CIG project AutoGameDesign (project no: 630665).

\ifCLASSOPTIONcaptionsoff
  \newpage
\fi

\bibliographystyle{IEEEtran}
\bibliography{surprise_jrnl}




\end{document}